%% file: main.tex
\documentclass[runningheads]{llncs}

\usepackage[year=2026]{eccv}

\usepackage{eccvabbrv}

\usepackage{graphicx}
\usepackage{booktabs}
\usepackage{booktabs}     
\usepackage{multirow}    
\usepackage{graphicx}    
\usepackage{algorithm}
\usepackage{algorithmic}
\usepackage[table]{xcolor} 

\usepackage[accsupp]{axessibility}  %
\def\method{LESV}

\usepackage{hyperref}

\usepackage{orcidlink}

\usepackage{ulem}

\begin{document}

\title{LESV: Language Embedded Sparse Voxel Fusion for Open-Vocabulary 3D Scene Understanding}

\titlerunning{Abbreviated paper title}

\author{Fusang Wang\inst{1,2} \and
Nathan Piasco\inst{1} \and
Moussab Bennehar\inst{1} \and
Luis Rold\~ao\inst{1} \and
Dzmitry Tsishkou\inst{1} \and
Fabien Moutarde\inst{2}
}

\authorrunning{F. Wang et al.}

\institute{Huawei Noah's Ark Lab \and
CAOR, Mines Paris-PSL, France \\
}

\maketitle
\input{figs/teaser}
\input{secs/0_abstract}    
\input{secs/1_intro}
\input{secs/2_related_works}

\input{secs/3_method}
\input{secs/4_Experiments}
\input{secs/5_Conclusion}
\input{secs/X_suppl}
\bibliographystyle{splncs04}
\bibliography{main}
\end{document}

%% file: figs/teaser.tex
\begin{figure*}[ht]
    \vspace{-0.5 cm}
    \centering 
    \includegraphics[width=0.98\textwidth]{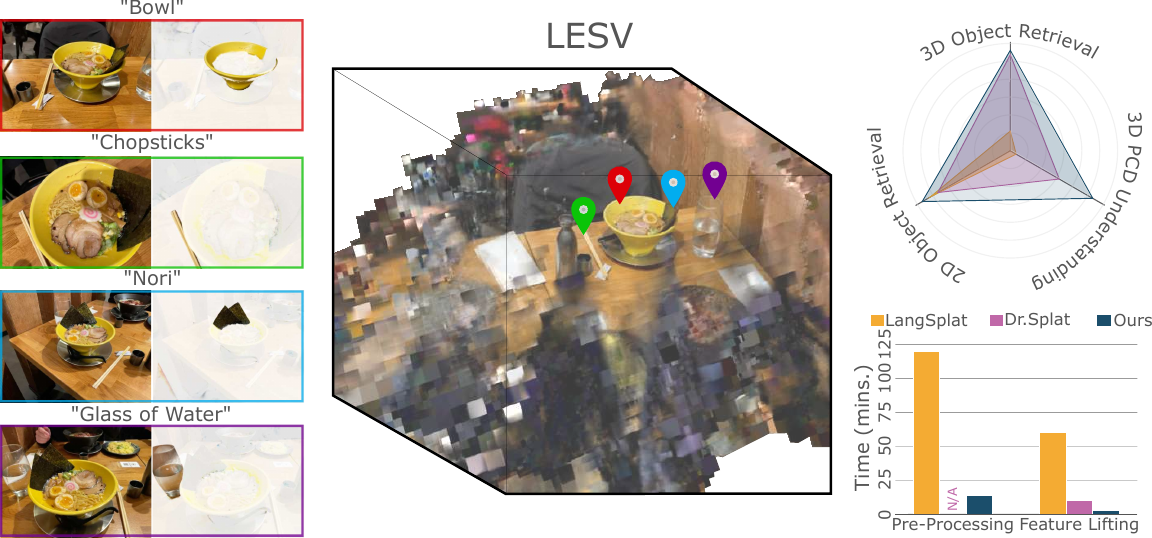} 
\caption{By registering dense, language-aligned features from the AM-RADIO foundation model onto an explicit Sparse Voxel Representation, \method~enables precise, deterministic localization of complex, fine-grained queries directly in 3D. This structured volume fusion facilitates general-purpose scene understanding, excelling in tasks such as open-vocabulary 3D object retrieval, 2D object localization, and point cloud understanding. Consequently, \method~establishes a new state-of-the-art across diverse 2D and 3D benchmarks, while drastically reducing feature lifting and data preprocessing time}
    \vspace{-0.5 cm}
    \label{fig:teaser}
\end{figure*}

%% file: secs/0_abstract.tex
\begin{abstract}
Recent advancements in open-vocabulary 3D scene understanding heavily rely on 3D Gaussian Splatting (3DGS) to register vision-language features into 3D space. However, we identify two critical limitations in these approaches: the \textit{spatial ambiguity} arising from unstructured, overlapping Gaussians which necessitates probabilistic feature registration, and the \textit{multi-level semantic ambiguity} caused by pooling features over object-level masks, which dilutes fine-grained details. 
To address these challenges, we present a novel framework that leverages Sparse Voxel Rasterization (SVRaster) as a structured, disjoint geometry representation. 
By regularizing SVRaster with monocular depth and normal priors, we establish a stable geometric foundation. This enables a deterministic, confidence-aware feature registration process and suppresses the semantic bleeding artifact common in 3DGS. Furthermore, we resolve multi-level ambiguity by exploiting the emerging dense alignment properties of the \textbf{AM-RADIO} foundation model, avoiding the computational overhead of hierarchical training methods.
Our approach achieves state-of-the-art performance on Open Vocabulary Point Cloud Understanding, and highly competitive results on 3D Object Retrieval benchmarks. 

\end{abstract}

%% file: secs/1_intro.tex
\section{Introduction}
The ability to query and interact with 3D environments using open-vocabulary natural language is a cornerstone of next-generation robotics and augmented reality~\cite{Alama2025RayFrontsOS, li2025scenesplat, halacheva2025gaussianvlm, ding2023pla, liu2023weakly}. While early approaches leveraged implicit Neural Radiance Fields (NeRFs)~\cite{2020NeRF}, the community has recently shifted toward explicit 3D Gaussian Splatting (3DGS)~\cite{Shi2023LanguageE3DGS} to enable real-time rendering of feature fields. Pioneering methods such as LangSplat \cite{Qin2023LangSplat3L, Li2025LangSplatV2H3} and LeGaussians~\cite{Shi2023LanguageE3DGS} view 3D scene understanding as a view-synthesis problem: they lift language embeddings like CLIP~\cite{radford2021learning} into 3D Gaussians by supervising rendered 2D feature maps against pre-extracted priors.
However, this ``render-and-compare'' paradigm is computationally expensive and decoupled from direct spatial reasoning in 3D~\cite{Wu2024OpenGaussianTP}. To address this, recent works~\cite{JunSeong2025DrSD, Chiou2026ProFuseEC} propose \textit{direct feature registration}. Instead of optimizing features via 2D supervision, these methods directly lift 2D vision-language features onto 3D primitives using inverse volume rendering. While this shift reduces the feature-lifting time from hours to minutes, we identify two fundamental limitations that prevent current registration pipelines from achieving high-fidelity retrieval.

\textbf{Spatial Ambiguity in Unstructured Geometry.} 3DGS represents scenes using a collection of unstructured, anisotropic Gaussians that significantly overlap. Consequently, determining which Gaussian should be assigned a specific high-dimensional feature becomes a complex, heuristic-driven task~\cite{JunSeong2025DrSD}. This inherent ambiguity results in semantic bleeding and ray-like spillovers at the object boundaries~\cite{Chiou2026ProFuseEC}, preventing the precise delineation required for point-level 3D interaction.

\textbf{Multi-Level Semantic Ambiguity.} A single point in 3D space often resides within multiple semantic hierarchies simultaneously, a challenge known as \textit{point ambiguity}~\cite{Qin2023LangSplat3L}. For instance, a spatial coordinate on a bear's nose should ideally respond to queries for its sub-part (`nose'), parent component (`head'), and global category (`bear'). A widely adopted solution is to resolve this by leveraging Segment Anything Model (SAM)~\cite{kirillov2023segment} hierarchies to train multiple independent feature fields~\cite{Qin2023LangSplat3L}. While effective, such an approach imposes massive pre-processing overhead, given redundant optimization and registration passes for every semantic hierarchy. 

To resolve the aforementioned challenges, we propose a paradigm shift in both scene representation and feature encoding for 3D language fields. For spatial ambiguity, we move beyond the probabilistic nature of Gaussians by adopting Sparse Voxel Rasterization (SVRaster)~\cite{Sun2024SparseVR}. 
This leverages both an explicit voxel geometry and a well-defined surface, enabling deterministic feature registration and avoiding severe memory bottlenecks. For multi-level semantic ambiguity, we build upon the emergent properties of Agglomerative Vision Foundation Model, specifically AM-RADIO~\cite{Ranzinger2023AMRADIOAV}. Unlike CLIP's global alignment, AM-RADIO is trained to distill both fine-grained spatial signals from DINOv2~\cite{oquab2023dinov2}, SAM~\cite{kirillov2023segment} and high-level language semantics from SigLIP~\cite{Zhai2023SigLip}, effectively allowing spatial tokens to inherit global content while retaining local semantics \cite{Alama2025RayFrontsOS}. By aligning the spatial token with the language feature space, we generate a semantic field that naturally supports multi-scale queries in a single pass.
%
Our contributions can be summarized as follows:
\begin{itemize}
\item We exploit Sparse Voxel Rasterization (SVRaster)\cite{Sun2024SparseVR} for 3D feature fields. Its explicit, disjoint voxel geometry, facilitates deterministic feature registration, circumventing the memory bottlenecks from inverse volume rendering.
\item By relying on agglomerative vision foundation models~\cite{Ranzinger2023AMRADIOAV}, we resolve the multi-level ambiguity inherent in hierarchical methods. This enables precise and fine-grained querying while drastically reducing feature preprocessing time. 
\item Through extensive experiments on 2D and 3D benchmarks, we demonstrate that \method~establishes a new state-of-the-art in open-vocabulary point cloud understanding~\cite{Dai2017ScanNetR3} and achieves competitive performance on 3D object retrieval and 2D localization~\cite{Kerr2023LERFLE}.
\end{itemize}

%% file: secs/2_related_works.tex
\section{Related Work}

\subsection{Language-Embedded 3D Scene understanding}
The integration of vision-language priors~\cite{Radford2021CLIP, Zhai2023SigLip} into 3D reconstruction has evolved from implicit neural representations~\cite{2020NeRF} to explicit, real-time primitives like 3D Gaussian Splatting (3DGS)~\cite{Kerbl20233DGS, Shi2023LanguageE3DGS}. Early approaches such as LERF~\cite{Kerr2023LERFLE} grounded CLIP features within Neural Radiance Fields (NeRFs). However, relying on volumetric ray-marching to query language embeddings is computationally expensive and requires multi-scale rendering during inference to handle different semantic levels~\cite{Qin2023LangSplat3L}. To address the computational bottlenecks of NeRF-based language fields, recent research~\cite{Shi2023LanguageE3DGS, Qin2023LangSplat3L, shen2025trace3d, Zhou2023Feature3S} has pivoted toward 3DGS. These frameworks optimize 3D feature fields by explicitly supervising rendered 2D features against pre-computed feature maps. Because directly splatting these high-dimensional feature vectors (e.g., 512-D for CLIP) incurs prohibitive memory costs and precludes real-time performance, existing frameworks employ dimensionality reduction strategies. For instance, LangSplat~\cite{Qin2023LangSplat3L} and Feature3DGS~\cite{Zhou2023Feature3S} compress features into a low-dimensional latent space via scene-specific autoencoders.
More recently, voxel-based representations have begun to emerge as a promising alternative backbone representation for language field construction~\cite{huang2026openvoxel}, motivating our adoption of Sparse Voxel Rasterization as a structured foundation for feature lifting.

\subsection{Direct Feature Registration with 3DGS}

While feature distillation has proven effective for novel view synthesis, it is poorly suited to direct 3D queries: features stored in individual Gaussians are typically only meaningful after alpha-blending in 2D, biasing the representation toward rendering rather than 3D understanding~\cite{JunSeong2025DrSD, Wu2024OpenGaussianTP, arafa2025beyond}. To address this, recent methods pursue different strategies. Some approaches leverage multi-view consensus during feature distillation~\cite{arafa2025beyond}, others employ contrastive learning with SAM masks to group multi-view 2D semantics into coherent 3D objects~\cite{Wu2024OpenGaussianTP, cen2025tackling}. However, these approaches still require an additional optimization stage after the 3D representation is trained. 
An alternative is direct feature registration, which lifts 2D features onto 3D Gaussians via inverse volume rendering~\cite{JunSeong2025DrSD, Chiou2026ProFuseEC, cheng2024occam}. 
For example, Dr.~Splat~\cite{JunSeong2025DrSD} registers embeddings by mapping per-pixel CLIP features onto the top-$K$ Gaussians intersected by each viewing ray, enabling more 3D-native retrieval. 
Despite the fast lifting speed and improved 3D-native precision, the registration remains probabilistic and ray-dependent, which can dilute semantics and produce spillover false positives during 3D queries~\cite{Chiou2026ProFuseEC}. ProFuse~\cite{Chiou2026ProFuseEC} mitigates these artifacts by aggregating 2D masks into 3D context proposals prior to registration and concentrating feature assignments onto a small set of Gaussians. However, the fundamental ambiguity induced by alpha-blending is not fully resolved. This motivates the use of structured geometric backbones, which support more deterministic and decoupled feature mapping.

%% file: secs/3_method.tex
\section{Method}


\subsection{Preliminary: Sparse Voxel Rasterization (SVRaster)}

Unlike the unstructured ellipsoids used in 3D Gaussian Splatting (3DGS), SVRaster represents the scene as a set of sparse voxels $\mathcal{V} = \{v_1, \dots, v_N\}$ organized within a high-resolution hierarchical grid~\cite{Sun2024SparseVR}. Each voxel $v_i$ is parameterized by its spatial center $\mathbf{x}_i \in \mathbb{R}^3$, a set of density values defined at its eight corners $\boldsymbol{\sigma}_i = \{\sigma_0, \dots, \sigma_7\}$, and a view-dependent color represented by Spherical Harmonic (SH) coefficients. 

To render the scene, SVRaster projects voxels onto the image plane and sorts them using View-Dependent Morton Codes. This strict depth ordering explicitly prevents the popping artifacts prevalent in unstructured 3DGS architectures. The final pixel color $C$ and depth $D$ are computed by accumulating the contributions of sorted voxels $\mathcal{N}$ intersecting the pixel ray:
\begin{equation}
    C = \sum_{j \in \mathcal{N}} T_j \alpha_j \mathbf{c}_j, \quad D = \sum_{j \in \mathcal{N}} T_j \alpha_j z_j, \quad \text{with} \quad T_j = \prod_{m=1}^{j-1} (1 - \alpha_m),
\end{equation}
where $z_j$ is the distance from the camera center to the $j$-th voxel center. The opacity $\alpha_j$ is derived from the volume density $\sigma(\mathbf{x}_j)$ and the ray-voxel intersection interval $\delta_j$ as $\alpha_j = 1 - \exp(-\sigma(\mathbf{x}_j) \delta_j)$. The density $\sigma(\mathbf{x}_j)$ is computed via trilinear interpolation of the corner densities $\boldsymbol{\sigma}_j$, ensuring a continuous density field within the explicit grid .

\paragraph{Base Volume Fusion.} The structured voxel representation of SVRaster facilitates a deterministic mapping between 2D pixels and 3D primitives. For a given voxel $v_i$, a robust global feature $\mathbf{F}_i$ can be aggregated from multi-view 2D feature maps $\mathcal{F}_k$ via weighted volume fusion:
\begin{equation}
    \mathbf{F}_i = \frac{\sum_{k \in \Omega_i} w_{i,k} \mathbf{f}_{i,k}}{\sum_{k \in \Omega_i} w_{i,k} + \epsilon}, \quad \text{where} \quad \mathbf{f}_{i,k} = \text{Bilinear}(\mathcal{F}_k, \pi_k(\mathbf{x}_i)).
    \label{equ:volume fusion}
\end{equation}
Here, $\Omega_i$ denotes the set of visible views and $\pi_k(\cdot)$ is the camera projection. To ensure geometric adherence, the weight $w_{i,k}$ is defined by a Gaussian kernel based on the depth discrepancy: $w_{i,k} = \exp( - |z_{i,k} - \mathbf{D}_k|^2 / 2\beta^2 )$, which prioritizes features sampled near the rendered surface $\mathbf{D}_k$. The term $\beta$ represents the bandwidth of the Gaussian kernel, which explicitly controls the sensitivity of the fusion weights to spatial depth discrepancies.




\subsection{Geometric-Aware Robust Volume Fusion}

\input{figs/method_overview}



\subsubsection{Geometric Refinement via Scene Constraints.}
While the explicit grid structure of SVRaster conceptually simplifies the lifting of high-dimensional features, achieving high-fidelity 3D feature fields remains non-trivial. The vanilla SVRaster optimization prioritizes photometric fidelity over structural integrity, frequently resulting in fragmented or "hollow" underlying geometry. These structural impurities produce noisy surface renderings that severely disrupt the volumetric fusion. To optimize the underlying explicit geometry of SVRaster, we incorporate 2D monocular priors into the training objective.

\paragraph{Patch-wise Depth Consistency.} 
Directly penalizing absolute depth differences between multi-view renderings and monocular priors often destabilizes optimization due to inherent global scale and shift ambiguities. To mitigate this without compromising local topology, we strategically adopt a patch-level structural constraint, $\mathcal{L}_{\text{patch}}$, following \cite{Li2024DNGaussianOS}. 
For each localized patch $p \in \mathcal{P}$, the loss is formulated as:
\begin{equation}
    \mathcal{L}_{\text{patch}} = \frac{1}{|\mathcal{P}|} \sum_{p \in \mathcal{P}} \left\| \left( \frac{\mathbf{D}_{\text{ren}}(p) - \mu_{\text{ren}}}{\sigma_{\text{ren}}} \right) - \left( \frac{\mathbf{D}_{\text{prior}}(p) - \mu_{\text{prior}}}{\sigma_{\text{prior}}} \right) \right\|^2_2,
\end{equation}
where $\mu$ and $\sigma$ denote the mean and standard deviation of depth values within patch $p$. This normalized formulation is advantageous to inherit high-fidelity relative structures like edges and local curvature from the monocular prior .

\paragraph{Normal Consistency Supervision.} 
While patch-wise depth regularizes zero-order spatial positioning, voxel-based feature aggregation remains highly sensitive to first-order surface variations, particularly at grazing angles and object boundaries where semantic leakage is most severe. To suppress these artifacts, we incorporate an analytic normal constraint, $\mathcal{L}_{\text{norm}}$, following \cite{Yu2022MonoSDFEM}:
\begin{equation}
    \mathcal{L}_{\text{norm}} = \frac{1}{|\mathcal{P}|} \sum_{p \in \mathcal{P}} \left( 1 - \mathbf{N}_{\text{ren}}(p) \cdot \mathbf{N}_{\text{prior}}(p) \right),
\end{equation}
where $(\cdot)$ denotes the vector dot product.

\subsubsection{Robust Volume Fusion via Multi-Level TSDF Confidence.}

Despite monocular geometric regularization, the rendered depths in SVRaster can still exhibit view-dependent inconsistencies. Since the fusion weight uses an exponential kernel, errors in the rendered depth $\mathbf{D}_{\text{ren},k}$ directly distort the aggregation weights. 
To resolve this, we leverage an explicit voxel grid advantage: extracting a 3D-consistent depth prior from the underlying Truncated Signed Distance Function (TSDF) field within the grid. This can be exploited to compute an inherently consistent mesh, naturally aligned with the global 3D scene representation and free from single-view rendering noise.

Standard mesh extraction operating on a single resolution tier typically forces a sub-optimal geometric trade-off: extraction from coarse voxels removes high-frequency structural details, whereas fine voxels yield incomplete geometries riddled with holes due to sparsely populated or unobserved TSDF values. We mitigate this by fusing TSDF information across multiple resolution levels, producing a mesh that preserves fine topological details without compromising global surface continuity
(cf. supplementary).
We then project this multi-level mesh into each view $k$ to obtain a reference depth $\mathbf{D}_{\text{mesh},k}$
and formulate a continuous geometric confidence map, $w_{\text{conf}, k}$, defined by the absolute depth discrepancy:
\begin{equation}
    w_{\text{conf}, k} = \exp\left(-\frac{|\mathbf{D}_{\text{mesh}, k} - \mathbf{D}_{\text{ren}, k}|}{2\sigma}\right),
\end{equation}
where $\sigma$ controls the decay. This confidence establishes a principled trade-off between spatial proximity and geometric accuracy.
Finally, we inject this confidence into the volumetric fusion process by dynamically modulating the original spatial weight $w_{s,i,k}$ from Eq.~\ref{equ:volume fusion} and replacing it by our robust aggregation weight used to aggregate the 3D feature $\mathbf{F}_i$ across all visible views:
\begin{equation}
    w_{i, k} = w_{s, i, k} \cdot w_{\text{conf},i, k}, 
\end{equation}

\subsubsection{Scalable Batch Fusion.}

Direct feature registration in 3D Gaussian Splatting (3DGS) architectures faces severe memory bottlenecks (e.g., Dr.~Splat~\cite{JunSeong2025DrSD} requires over 256GB of RAM for dense scenes). Because 3DGS defines surfaces implicitly via volumetric accumulation, aggregating 2D features onto 3D primitives necessitates global depth-sorting of all Gaussians to compute view-dependent alpha-blending weights. In contrast, the explicit grid structure of SVRaster inherently resolves this bottleneck. As demonstrated in our fusion formulation, the spatial distance from any voxel to the physical surface is deterministic. Therefore, our volume fusion process computes feature weights based solely on local geometric proximity and confidence, entirely bypassing the need for view-dependent global sorting. This spatial independence ensures voxels remain strictly disjoint during fusion. Consequently, we can partition the grid and execute multi-view feature projection in isolated spatial batches, enabling highly scalable, $\mathcal{O}(1)$ memory processing regardless of scene scale.


\subsection{Resolving Multi-Level Ambiguity via Agglomerative Features}

\subsubsection{Emergent Dense Alignment in AM-RADIO.}
Rather than relying on the standard SAM+CLIP pipeline, we adopt AM-RADIO~\cite{Ranzinger2023AMRADIOAV} as our feature backbone, which distills multiple teacher signals — CLIP~\cite{Radford2021CLIP}, SAM~\cite{kirillov2023segment}, and DINOv2~\cite{oquab2023dinov2} — into a single unified backbone. Given an input image $I$, the RADIO backbone $\Phi_{\text{RADIO}}$ produces a sequence of tokens consisting of a global class token $\mathbf{t}_{cls}$ and a set of spatial patch tokens $\mathbf{T}_{patch} \in \mathbb{R}^{N \times D}$:
\begin{equation}
    [\mathbf{t}_{cls}, \mathbf{T}_{patch}] = \Phi_{\text{RADIO}}(I).
\end{equation}
In standard vision-language alignment, a global semantic vector is obtained by projecting the CLS token through a specific language-aligned MLP head, denoted as $h_{\text{lang}}(\cdot)$. However, recent works have observed an emergent property in agglomerative models: the dense patch tokens preserve strong alignment with the language space even without explicit dense supervision~\cite{Alama2025RayFrontsOS}. 
Instead of relying solely on the global $\mathbf{t}_{cls}$, we directly project the spatial patch tokens through the language head MLP to generate a dense, language-aligned feature map $\mathbf{F}_{\text{lang}} \in \mathbb{R}^{H \times W \times C}$ :
\begin{equation}
    \mathbf{F}_{\text{lang}} = h_{\text{lang}}(\mathbf{T}_{patch}).
\end{equation}
By utilizing these patch tokens, which inherently encode both local texture and global semantics, $\mathbf{F}_{\text{lang}}$ allows the generated feature field to simultaneously capture multi-scale concepts without the computational overhead of explicit hierarchical SAM-based mask pooling.

Despite these advantages, the direct application of patch-level language features introduces spatial resolution bottlenecks, particularly when segmenting small objects. Because the AM-RADIO architecture is optimized for a native input resolution of $512 \times 512$ (with a strict upper bound of 4096), naively upscaling the input image to recover spatial granularity is architecturally sub-optimal and computationally prohibitive.

\subsubsection{Self-Correcting Sliding Window Upsampling.}

To circumvent these architectural constraints and extract dense semantic features at the native image resolution, we introduce a self-correcting sliding window upsampling strategy. 
While learning-based deep upsampling architectures (e.g., AnyUp \cite{Wimmer2025AnyUpUF}) can arbitrarily increase spatial resolution, empirical evaluations indicate that their attention-based spatial mixing heavily distorts the strict high-dimensional semantic space required for zero-shot 3D retrieval 
(cf. supplementary).

To bypass this semantic degradation and maintain strict feature-space fidelity, we adopt a sliding window inference approach combined with spatial Gaussian attenuation. We partition the high-resolution input into overlapping crops $\{C_1, \dots, C_M\}$. To reconstruct a continuous, high-resolution feature map $\mathbf{F}_{\text{high}}$ while explicitly mitigating hard boundary discontinuities between adjacent crops, we aggregate the localized predictions using a Gaussian kernel. For a given pixel coordinate $\mathbf{u}$ in the full-resolution image coordinate frame, the aggregated feature is defined as:
\begin{equation}
    \mathbf{F}_{\text{high}}(\mathbf{u}) = \frac{\sum_{k=1}^{M} G(\mathbf{u} - \mathbf{c}_k) \cdot \mathbf{F}_k(\mathbf{u} - \mathbf{c}_k)}{\sum_{k=1}^{M} G(\mathbf{u} - \mathbf{c}_k)}
\end{equation}
where $\mathbf{c}_k$ denotes the top-left spatial anchor of crop $k$, $\mathbf{F}_k$ represents the bilinearly upsampled AM-RADIO feature map for the crop, and $G(\cdot)$ is a 2D Gaussian kernel that assigns smoothly decaying weights near the crop boundaries.

\paragraph{Self-Correlating Aggregation.}
Although Gaussian weighting smoothly blends overlapping regions, the raw dense features extracted from AM-RADIO often retain high-frequency noise. To resolve this, we integrate the Self-Correction Recursive Attention (SCRA) and Self-Correlating Global Aggregation (SCGA) mechanisms from RADSeg~\cite{Alama2025RADSegUP}. These modules replace standard dot-product attention with a thresholded cosine-similarity formulation that explicitly masks out negatively correlated token pairs. By acting as a strict semantic filter, this hard-thresholding approach effectively eliminates the discordant noise injected during the window fusion process. Consequently, the final dense feature map, $\mathbf{F}_{\text{high}}$, retains semantic fidelity and sharp object boundaries prior to integration into our 3D volumetric fusion pipeline.

%% file: figs/method_overview.tex
\begin{figure*}[t]
    \centering    
    \includegraphics[width=\linewidth]{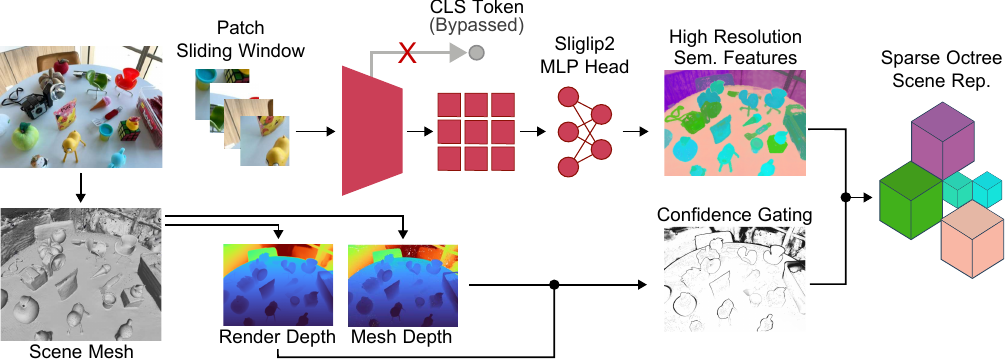}
    \caption{\textbf{Overview of the \method~Architecture.} \textbf{Top:} Input images are processed via a sliding window through AM-RADIO and a SigLIP-2 MLP to extract high-resolution language aligned features. \textbf{Bottom:} Comparing rendered depth against explicit \textit{SVRaster} mesh depth generates a geometric confidence map, which selectively gates the feature projection, ensuring high-fidelity volume fusion (\textbf{Right}).}
    \label{fig:method_overview}
\end{figure*}

%% file: secs/4_Experiments.tex
\section{Experiments}

\subsection{Implementation Details.}
All experiments and performance analyses were conducted on a single workstation equipped with a high-end GPU (featuring compute and VRAM capabilities equivalent to an NVIDIA RTX 5090). Notably, because the original open-source implementation of Dr.~Splat~\cite{JunSeong2025DrSD} requires prohibitive system memory allocations (exceeding 256GB of RAM on dense scenes), we integrated optimized memory management routines to evaluate it locally. This allowed us to successfully replicate their officially reported baseline metrics within standard hardware limits, ensuring a rigorously fair comparison.

\subsection{Open-Vocabulary Object Retrieval}
\input{figs/multi_level}

\paragraph{Evaluation Protocol and Metrics.}
To assess our model's spatial querying capabilities, we evaluate open-vocabulary object retrieval on the LERF~\cite{Kerr2023LERFLE} dataset across both 3D and 2D settings. The fundamental distinction between these two tasks lies in the domain where semantic thresholding is applied. For \textbf{3D object retrieval}, we follow the evaluation protocol of Dr.~Splat~\cite{JunSeong2025DrSD}, where all methods are evaluated under a single object-level SAM mask setting. Objects are retrieved by thresholding cosine similarities directly in 3D space, without any mean filtering applied to the rendered masks during metric calculation. Since AM-RADIO features exhibit a different value distribution compared to standard SAM+CLIP paradigms, we adopt normalized cosine similarity for evaluated methods to align feature spaces prior to 3D thresholding, ensuring a fair and consistent comparison across methods.
For \textbf{2D object retrieval}, we follow the evaluation strategy of LangSplat~\cite{Qin2023LangSplat3L}, which first renders the relevancy map onto the 2D image plane from a specified camera viewpoint, and subsequently applies thresholding and mean filtering in 2D space to yield the final results.

\paragraph{3D Object Retrieval.} 
Quantitative results are presented in Table~\ref{tab:lerf_results}. \method~achieves competitive state-of-the-art results on this benchmark, attaining the second-best average mIoU (56.11\%) and the best Acc@25 (85.21\%), outperforming the registration-based baseline Dr.~Splat. Notably, dense language features from AM-RADIO effectively resolve multi-level semantic ambiguities, yielding substantial improvements on scenes dominated by complex sub-object queries (e.g., the \textit{ramen} scene). Figure~\ref{fig:multi_level} further qualitatively illustrates our method's superiority in handling multi-level linguistic ambiguities. We additionally report LaGa~\cite{cen2025tackling} under a different evaluation setting (marked $^{\dagger}$) using multi-level SAM masks, as hierarchical contrastive learning is a core contribution of the work. Although LaGa~\cite{cen2025tackling} achieves the best mIoU under this setting, this comes at a significantly higher preprocessing and training cost ($\sim$190 mins vs. $\sim$20 mins for \method).
\input{tabs/lerf_ovs}

Figure~\ref{fig:lerf_viz} demonstrates that our method effectively eliminates the spillover artifacts prevalent in 3DGS, as evidenced by the ``plate'' and ``sink'' queries. The integration of dense AM-RADIO features robustly resolves multi-level ambiguities, successfully isolating fine-grained sub-parts, such as the ``corn'' within the ramen. Furthermore, we achieve precise segmentation of extremely small objects, such as the ``pirate hat'', which the baseline struggles to capture. Interestingly, we observe a highly nuanced semantic behavior enabled by dense feature formulation: when querying the ``refrigerator'', our methods segments the appliance's surface while excluding the unrelated photos attached to its door. Although this strict semantic decoupling is penalized by the coarse ground-truth annotations of the LERF benchmark, it highlights a fine-grained understanding of scene enabled by the method.

\input{tabs/2d_lerf}
\paragraph{Generalization to 2D Localization.} 
While distillation-based methods like LangSplat fail dramatically on the aforementioned pure 3D object retrieval task (averaging only 10.97\% mIoU), deterministic 3D registration provides a much more robust mechanism for scene understanding that naturally generalizes back to 2D projection tasks. As shown in Table~\ref{tab:lerf_2d_results_updated}, despite not being optimized or distilled for this setting, \method~achieves localization and retrieval performance on par with dedicated 2D methods such as LangSplat-v2~\cite{Li2025LangSplatV2H3} and Occam's LGS~\cite{cheng2024occam}, strictly matching the average localization accuracy of LangSplat-v2 (84.12\% mAcc vs. 84.10\% mAcc) and outperforming it on complex hierarchical scenes like \textit{ramen} (63.84\% mIoU vs. 51.87\% mIoU). Although Occam's LGS and LangSplat-v2 attain higher overall retrieval mIoU (61.28\% and 59.88\%, respectively), we emphasize that these methods are explicitly trained for the 2D setting, whereas \method~achieves comparable performance through generalization from its 3D-native feature lifting.

\input{figs/qualitative_fig}

\subsection{Open-Vocabulary Point Cloud Understanding}
\input{tabs/scannet_pcd}
\input{figs/scannet_viz}

\paragraph{Evaluation Protocol.} 
We evaluate open-vocabulary point cloud understanding on the ScanNet~\cite{Dai2017ScanNetR3} dataset across 19, 15, and 10-class configurations. To fairly evaluate distinct representations, we adapt the feature-transfer protocol from ProFuse~\cite{Chiou2026ProFuseEC}. For 3DGS baselines, we candidate nearest $K=64$ Gaussians shortlisted by Euclidean distance and weighted by distance weighting, formulated as $\exp(-\frac{1}{2} d^2)$, and Gaussian opacity. We omit Mahalanobis gating, as we empirically observed that it degrades performance in scenes dominated by small Gaussians. For \textit{SVRaster}, we simplify to a direct aggregation of the $K=8$ nearest voxel features using the identical distance weighting. 

\paragraph{Quantitative Results.} 
Quantitative evaluations are summarized in Table~\ref{tab:scannet_pcd_results}. \method~establishes a new SoTA, significantly outperforming the leading 3DGS-based baseline, Dr.~Splat. Specifically, on the highly challenging 19-class split, our method achieves a +21.56\% absolute improvement in mIoU (53.22\% vs. 31.66\%) and a +21.77\% leap in mAcc (70.41\% vs. 48.64\%). Furthermore, this substantial performance gap remains consistent across both the 15-class (+17.19\% mIoU) and 10-class (+20.38\% mIoU) settings. 

\paragraph{Qualitative Analysis.}
To contextualize these quantitative gains, we provide visual comparisons of the feature-transferred ScanNet point clouds in Figure~\ref{fig:scannet_pca_viz}. Qualitatively, \method~exhibits significantly reduced "semantic bleeding'' at object intersections compared to baseline methods and preserves more accurate spatial boundaries across varying levels of semantic granularity.

\subsection{Ablation Study}
To validate our architectural design, we conduct an ablation study across two complementary axes. In Table~\ref{tab:arch_ablation}, we isolate the geometric contribution by fixing both \method~and Dr.~Splat to identical SAM+CLIP features. The baseline \textit{SVRaster} yields sub-optimal geometry. Introducing patch-wise depth and surface normal supervision substantially refines structural fidelity. Integrating multi-level TSDF depth confidence further stabilizes results by down-weighting view-inconsistent projections, pushing \method~to outperform Dr.~Splat~\cite{JunSeong2025DrSD} on ScanNet (+0.40 mIoU, +2.03 mAcc). This validates that both the monocular cue integration and our proposed depth confidence gating are crucial components in making Sparse Voxels outperform the 3DGS counterpart.

In Table \ref{tab:feature_ablation}, we disentangle the effect of the feature backbone by evaluating all combinations of representation and encoder. Upgrading Dr. Splat from SAM+CLIP to SAM+AM-RADIO yields only marginal gains, as its 3DGS registration cannot scale to dense per-pixel features, forcing spatially rich AM-RADIO tokens into mask-level pooled embeddings and destroying fine-grained local semantics. In contrast, LESV's deterministic fusion directly registers dense patch-level features per voxel without mask pooling, fully preserving AM-RADIO's spatial richness. This confirms that the  performance gains arise from our dense feature registration paradigm rather than the encoder choice alone.
\input{tabs/arch_ablation}
\input{tabs/feature_ablation}

\subsection{Computational Efficiency}
To demonstrate practical scalability, we compare peak memory usage and computational overhead in Tables~\ref{tab:memory} and~\ref{tab:speed_analysis}, respectively. Dr.~Splat's official implementation exceeds 256\,GB RAM on dense scenes, revealing a structural limitation: mask-level proposal caching causes memory to scale non-linearly with image count. To obtain tractable baseline metrics, we optimized their implementation via chunking and FP16 precision with equivalent performance, resulting in the values reported in Table~\ref{tab:memory}. Scaling this paradigm to dense, per-pixel features inevitably causes catastrophic memory explosion, making dense feature registration infeasible for Dr.~Splat. In contrast, \method's explicit voxel grid and deterministic fusion enable spatial batching, bounding peak memory at $\sim$25\,GB ($\sim$14\,GB fusion, $\sim$11\,GB feature extraction) regardless of scene scale.
\input{tabs/peak_memory}
Regarding computational speed (Table~\ref{tab:speed_analysis}), optimization-based pipelines like LangSplat~\cite{Qin2023LangSplat3L} or LaGa~\cite{cen2025tackling} suffer from substantial preprocessing bottlenecks ($\sim$120 minutes per scene) due to the necessity of extracting multi-level SAM masks and CLIP features. OpenGaussian~\cite{Wu2024OpenGaussianTP} similarly follows this pipeline but limits extraction to single-level mask features, reducing overhead by $3\times$. Occam's LGS~\cite{cheng2024occam} proposes an efficient way to register features for 2D tasks, but it still necessitates multi-level registration and feature preprocessing. By leveraging dense features from a single forward pass of the AM-RADIO foundation model, \method~drastically reduces per-scene preprocessing time by over $8\times$ to approximately 14 minutes, while our deterministic fusion achieves a feature lifting time of merely $\sim$3 minutes. Dr.~Splat~\cite{JunSeong2025DrSD} bypasses per-scene preprocessing via a generalized pre-trained codebook built on the SAM+CLIP pipeline of LangSplat~\cite{Qin2023LangSplat3L}, and is therefore denoted ``-''.
\input{tabs/speed}

%% file: figs/multi_level.tex
\begin{figure*}[tb] 
    \centering
    \def\fgsize{0.24}
    \scriptsize
    \setlength{\tabcolsep}{0.0015\linewidth}
    \renewcommand{\arraystretch}{1.0}
    \begin{tabular}{cccc}
    
        & "bear" & "bear nose" & "bear legs" \\
        
        \rotatebox{90}{\hspace{10pt}Dr.Splat~\cite{JunSeong2025DrSD}} & 
        \includegraphics[clip=false, trim={0 0 0 0},width=\fgsize\columnwidth]{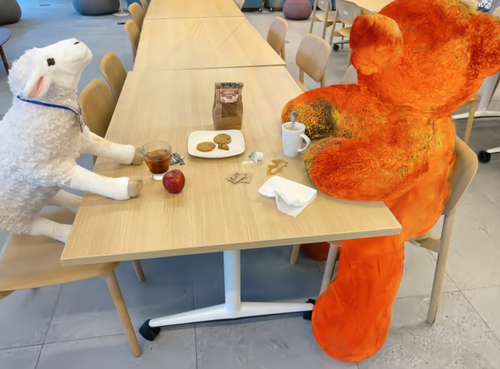} & 
        \includegraphics[clip=false, trim={0 0 0 0},width=\fgsize\columnwidth]{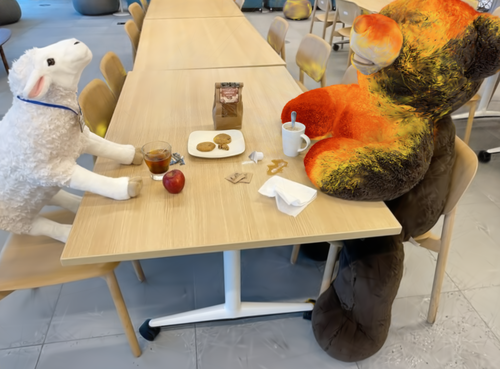} & 
        \includegraphics[clip=false, trim={0 0 0 0},width=\fgsize\columnwidth]{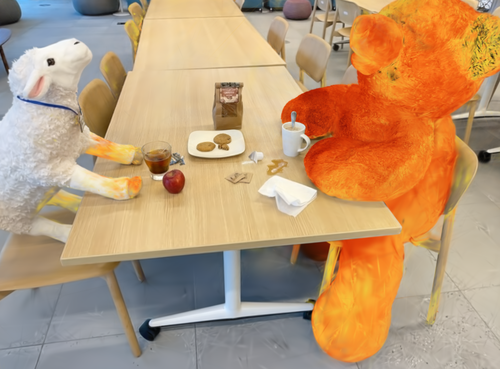} \\ 
        
        \rotatebox{90}{\hspace{20pt}Ours} & 
        \includegraphics[clip=false, trim={0 0 0 0},width=\fgsize\columnwidth]{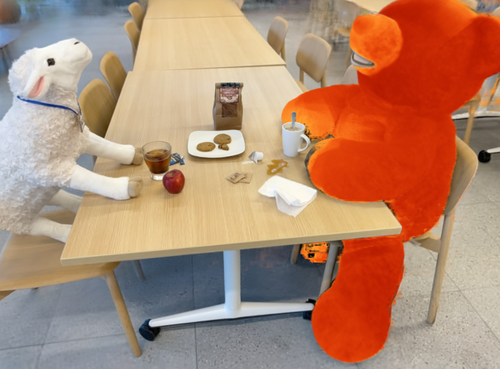} &
        \includegraphics[clip=false, trim={0 0 0 0},width=\fgsize\columnwidth]{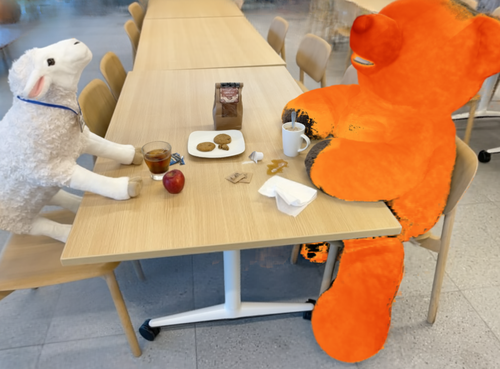} &
        \includegraphics[clip=false, trim={0 0 0 0},width=\fgsize\columnwidth]{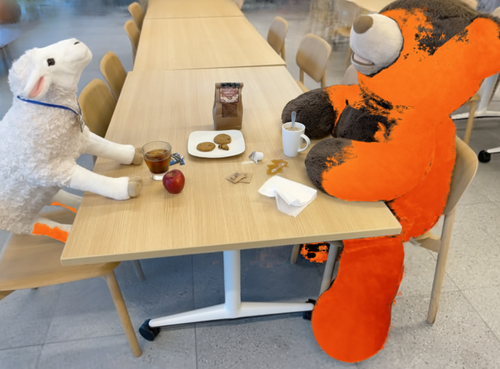} \\
        
    \end{tabular}
    \caption{\textbf{Multi-Level Semantic Ambiguity.} Comparison of querying on different semantic level, from a global ("bear") to  sub-parts ("bear nose", "bear legs"). Our method (bottom) dynamically concentrates semantic activation  onto the targeted regions.}
    \label{fig:multi_level}
\end{figure*}

%% file: tabs/lerf_ovs.tex
\begin{table*}[b]
    \centering
    \resizebox{0.9\linewidth}{!}{
    \begin{tabular}{l cc @{\hspace{9pt}} cc @{\hspace{9pt}} cc @{\hspace{9pt}} cc @{\hspace{9pt}} cc}
        \cmidrule(l{1pt}r{1pt}){2-11}
        & \multicolumn{2}{c}{{ramen}} & \multicolumn{2}{c}{{figurines}} & \multicolumn{2}{c}{{teatime}} & \multicolumn{2}{c}{{waldo\_kitchen}} & \multicolumn{2}{c}{\textbf{Average}} \\
        \cmidrule(l{2pt}r{2pt}){2-3} \cmidrule(l{2pt}r{2pt}){4-5} \cmidrule(l{2pt}r{2pt}){6-7} \cmidrule(l{2pt}r{2pt}){8-9} \cmidrule(l{2pt}r{2pt}){10-11}  
        \textbf{Method} & mIoU & Acc@25 & mIoU &Acc@25 & mIoU & Acc@25 & mIoU & Acc@25 & mIoU & Acc@25 \\
        \midrule
        LangSplat~\cite{Qin2023LangSplat3L} & 5.96 & 9.86 & 6.05 & 8.93 & 18.89 & 23.73 & 12.96 & 18.18 & 10.97 & 15.18 \\
        OpenGaussian~\cite{Wu2024OpenGaussianTP} & 31.01 & 42.25 & 39.29 & 55.36 & \underline{60.44} & \underline{76.27} & 22.70 & 31.82 & 38.36 & 51.43 \\
        Dr.~Splat~\cite{JunSeong2025DrSD} & \underline{37.49} & \underline{69.01} & \textbf{61.73} & \underline{83.93} & 59.45 & 72.88 & \textbf{52.00} & \textbf{81.81} & \underline{52.67} & \underline{76.91} \\     
        LaGa~\cite{cen2025tackling} & 31.60 & 46.48 & 49.01 & 83.93 & 58.73 & 89.83 & 38.80 & 63.64 & 44.54 & 70.97 \\
        \midrule
        LaGa$^{\dagger}$ (multi-level)~\cite{cen2025tackling} & 47.95 & 70.42 & 55.87 & 80.35 & 69.53 & 89.83 & 61.92 & 90.09 & 58.82 & 82.88 \\
        \midrule
        \method~(ours) & \textbf{53.34} & \textbf{81.69} & \underline{55.87} & \textbf{87.50} & \textbf{71.40} & \textbf{89.83} & \underline{43.84} & \textbf{81.81} & \textbf{56.11} & \textbf{85.21} \\
        \bottomrule
    \end{tabular}
    }
    \vspace{1mm}
    \caption{Open-vocabulary 3D object retrieval on the LERF dataset, reporting mIoU (\%) and Acc@25 (\%). \textbf{Bold} and \underline{underlined} denote the best and second-best results respectively. Results for LangSplat~\cite{Qin2023LangSplat3L} and LaGa~\cite{cen2025tackling} are re-implemented under the same evaluation protocol as Dr.~Splat~\cite{JunSeong2025DrSD}. LaGa$^{\dagger}$ (multi-level) is evaluated under a different setting using multi-level SAM masks and is included for reference.}
    \label{tab:lerf_results}
    \vspace{-10pt}
\end{table*}

%% file: tabs/2d_lerf.tex
\begin{table*}[h]
    \centering    
    \resizebox{0.9\linewidth}{!}{
    \begin{tabular}{l cc @{\hspace{9pt}} cc @{\hspace{9pt}} cc @{\hspace{9pt}} cc @{\hspace{9pt}} cc}
        \cmidrule(l{1pt}r{1pt}){2-11}
         & \multicolumn{2}{c}{ramen} & \multicolumn{2}{c}{figurines} & \multicolumn{2}{c}{teatime} & \multicolumn{2}{c}{waldo\_kitchen} & \multicolumn{2}{c}{\textbf{Average}} \\
        \cmidrule(l{2pt}r{2pt}){2-3} \cmidrule(l{2pt}r{2pt}){4-5} \cmidrule(l{2pt}r{2pt}){6-7} \cmidrule(l{2pt}r{2pt}){8-9} \cmidrule(l{2pt}r{2pt}){10-11} 
        \textbf{Method} & mIoU & Loc & mIoU & Loc & mIoU & Loc & mIoU & Loc & mIoU & Loc \\
        \midrule
        GAGS \cite{Peng2026GAGS} & 46.80 & 69.00 & 53.60 & 78.60 & 60.30 & \underline{88.10} & 55.80 & \underline{90.90} & 54.13 & 81.65 \\
        LangSplat \cite{Qin2023LangSplat3L} & 51.20 & 73.20 & 44.70 & 80.40 & 65.10 & \underline{88.10} & 44.50 & \textbf{95.50} & 51.38 & \textbf{84.30} \\
        LangSplatV2  \cite{Li2025LangSplatV2H3} & \underline{51.80} & \underline{74.70} & \underline{56.40} & \underline{82.10} & \underline{72.20} & \textbf{93.20} & \underline{59.10} & 86.40 & \underline{59.88} & 84.10 \\
        Occam’s LGS \cite{cheng2024occam} & 51.00 & 74.70 & \textbf{58.60} & 80.40 & 70.20 & \textbf{93.20} & \textbf{65.30} & 81.80 & \textbf{61.28} & 82.53 \\        
        \midrule
       \method~(ours) & \textbf{63.40} & \textbf{88.73} & 48.41 & \textbf{85.71} & 69.30 & 84.75 & 46.24 & 77.27 & 56.84 & \underline{84.12} \\
        \bottomrule
    \end{tabular}
    }
    \vspace{1mm}
    \caption{Open-vocabulary 2D object retrieval and localization results on the LERF dataset. We report mIoU (\%) for retrieval and Localization Accuracy (Loc \%). \textbf{Bold} indicates the best result, \underline{underline} indicates the second best. Note that the result is achieved without multi-level SAM mask generation but simple pass to Radio-v3}
    \label{tab:lerf_2d_results_updated}
    \vspace{-10pt}
\end{table*}

%% file: figs/qualitative_fig.tex
\begin{figure*}[tb] 
    \centering
    \def\fgsize{0.19}
    \scriptsize
    \setlength{\tabcolsep}{0.0015\linewidth}
    \renewcommand{\arraystretch}{1.0}
    \begin{tabular}{cccccc}%
        & "plate" & "corn" & "refrigerator" & "pirate hat" & "sink" \\ 
        
        \rotatebox{90}{\hspace{16pt}RGB} &
        \includegraphics[clip=false, trim={0 0 0 0},width=\fgsize\columnwidth]{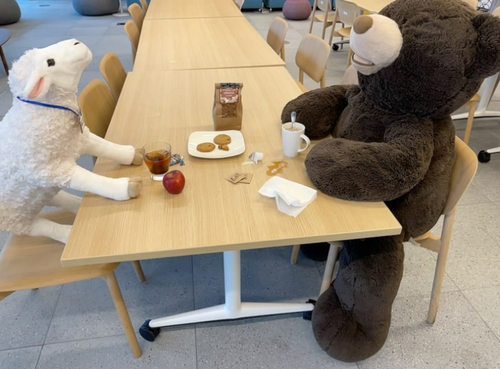} & 
        \includegraphics[clip=false, trim={0 0 0 0},width=\fgsize\columnwidth]{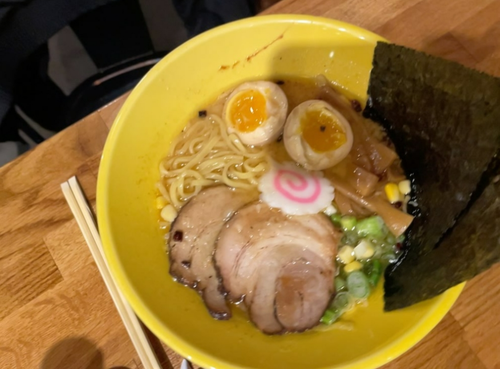} & 
        \includegraphics[clip=false, trim={0 0 0 0},width=\fgsize\columnwidth]{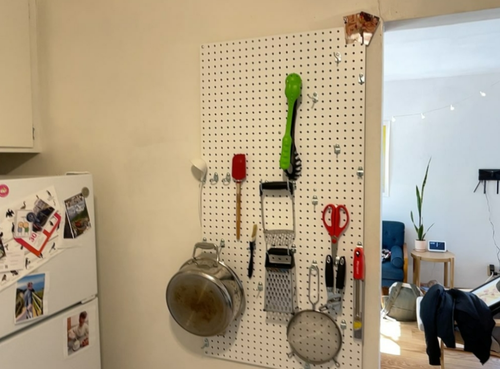} &  
        \includegraphics[clip=false, trim={0 0 0 0},width=\fgsize\columnwidth]{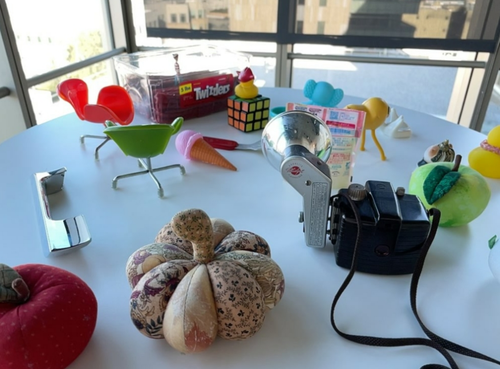} & 
        \includegraphics[clip=false, trim={0 0 0 0},width=\fgsize\columnwidth]{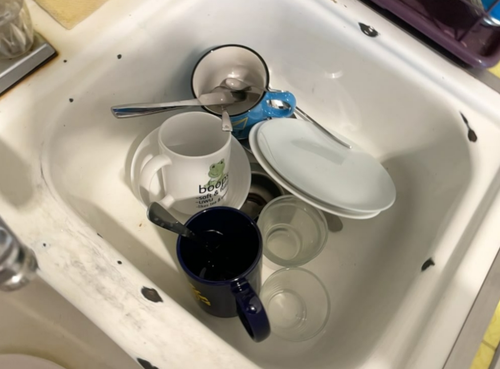} \\

        \rotatebox{90}{\hspace{6pt}Dr.Splat~\cite{JunSeong2025DrSD}} & 
        \includegraphics[clip=false, trim={0 0 0 0},width=\fgsize\columnwidth]{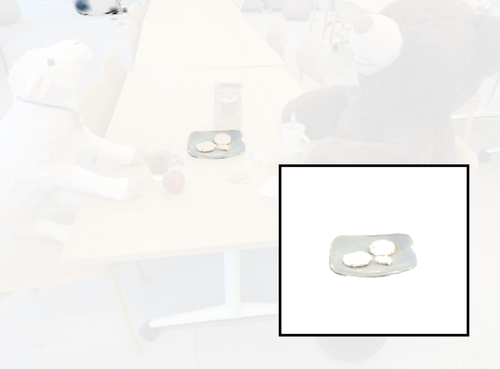} & 
        \includegraphics[clip=false, trim={0 0 0 0},width=\fgsize\columnwidth]{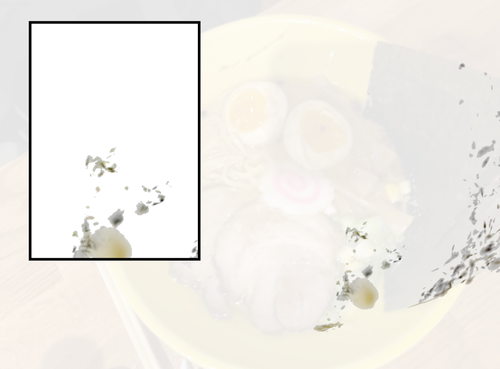} & 
        \includegraphics[clip=false, trim={0 0 0 0},width=\fgsize\columnwidth]{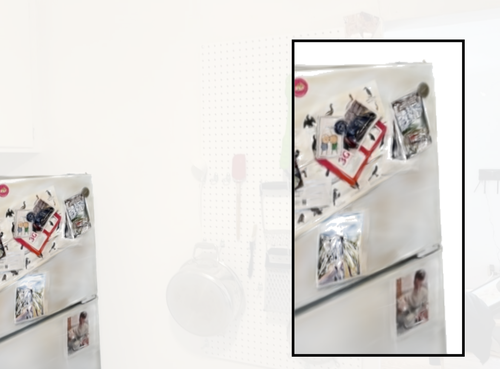} &  
        \includegraphics[clip=false, trim={0 0 0 0},width=\fgsize\columnwidth]{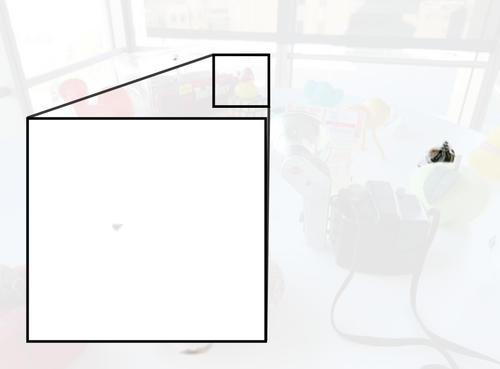} & 
        \includegraphics[clip=false, trim={0 0 0 0},width=\fgsize\columnwidth]{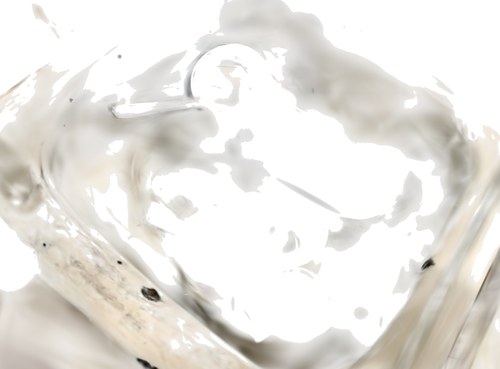} \\

        \rotatebox{90}{\hspace{16pt}Ours} & 
        \includegraphics[clip=false, trim={0 0 0 0},width=\fgsize\columnwidth]{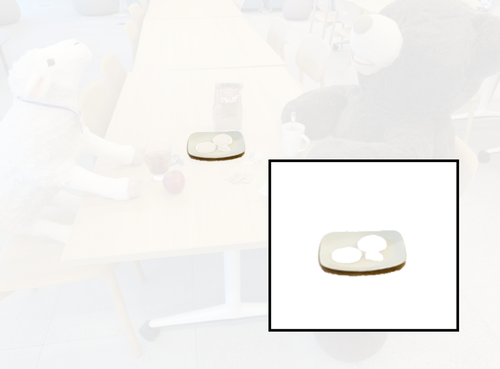} & 
        \includegraphics[clip=false, trim={0 0 0 0},width=\fgsize\columnwidth]{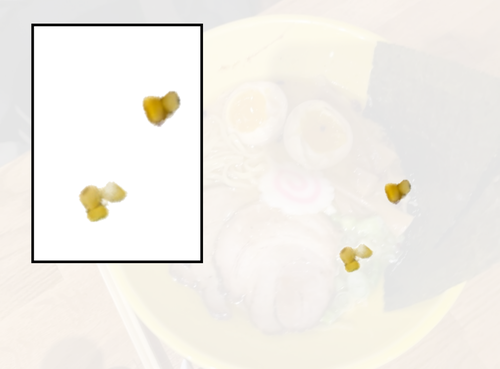} & 
        \includegraphics[clip=false, trim={0 0 0 0},width=\fgsize\columnwidth]{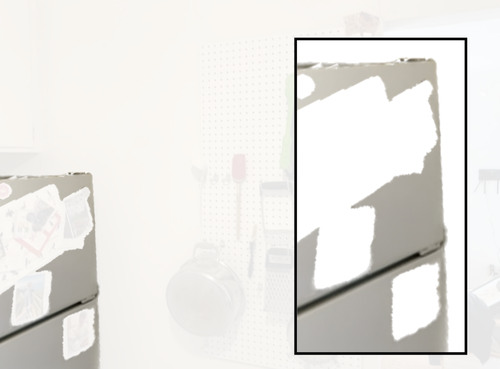} &  
        \includegraphics[clip=false, trim={0 0 0 0},width=\fgsize\columnwidth]{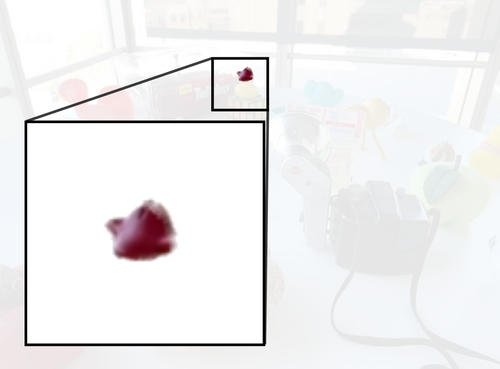} & 
        \includegraphics[clip=false, trim={0 0 0 0},width=\fgsize\columnwidth]{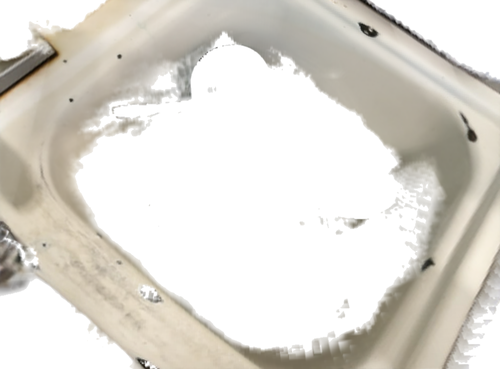} \\
    
    \end{tabular}
        \caption{\textbf{Qualitative Comparison of 3D Object Retrieval.} Visual results on the LERF dataset. Compared to the baseline Dr.~Splat, \method~effectively eliminates spillover artifacts ("plate", "sink"), precisely segments fine-grained sub-parts ("corn"), and captures extremely small objects ("pirate hat"). Notably, our method exhibits fine-grained semantic decoupling by accurately isolating the "refrigerator" surface while excluding attached photos.}
    \label{fig:lerf_viz}
\end{figure*}

%% file: tabs/scannet_pcd.tex
\begin{table*}[b]
    \centering
    \resizebox{0.6\linewidth}{!}{
    \begin{tabular}{l cc @{\hspace{9pt}} cc @{\hspace{9pt}} cc}
        \cmidrule(l{1pt}r{1pt}){2-7}
        & \multicolumn{2}{c}{\textbf{19 Classes}} & \multicolumn{2}{c}{\textbf{15 Classes}} & \multicolumn{2}{c}{\textbf{10 Classes}} \\
        \cmidrule(l{2pt}r{2pt}){2-3} \cmidrule(l{2pt}r{2pt}){4-5} \cmidrule(l{2pt}r{2pt}){6-7} 
        Method & mIoU & mAcc & mIoU & mAcc & mIoU & mAcc \\
        \midrule
        LangSplat \cite{Qin2023LangSplat3L} & 3.78 & 9.11 & 5.35 & 13.20 & 8.40 & 22.06 \\
        OpenGaussian \cite{Wu2024OpenGaussianTP} & 24.73 & 41.54 & 30.13 & 48.25 & 38.29 & 55.19 \\
        LaGa \cite{cen2025tackling} & \underline{32.50} & \underline{49.10} & 35.50 & 53.50 & 42.60 & 63.20 \\
        Dr.Splat \cite{JunSeong2025DrSD} & 31.66 & 48.64 & \underline{37.59} & \underline{56.54} & \underline{44.87} & \underline{64.27} \\
        \midrule
        \method~(ours) & \textbf{53.22} & \textbf{70.41} & \textbf{54.78} & \textbf{73.62} & \textbf{65.25} & \textbf{82.95} \\
        \bottomrule
    \end{tabular}
    }
\vspace{1mm}    
\caption{Open-vocabulary 3D semantic segmentation results on the ScanNet dataset across different class granularities (19, 15, and 10 classes)
\label{tab:scannet_pcd_results}}
\end{table*}

%% file: figs/scannet_viz.tex
\begin{figure*}[t]
    \centering
    \newcommand{\imgwidth}{0.23\textwidth}
    \newcommand{\imgheight}{2cm} 
    \scriptsize
    \renewcommand{\arraystretch}{0.5}
    \setlength{\tabcolsep}{1pt}
    
    \begin{tabular}{ccccc}
        & Scene~0000 & Scene~0140 & Scene~0590 & Scene~0645 \\
        
        \rotatebox{90}{\hspace{16pt}RGB GT} &
        \includegraphics[width=\imgwidth, height=\imgheight, keepaspectratio=false]{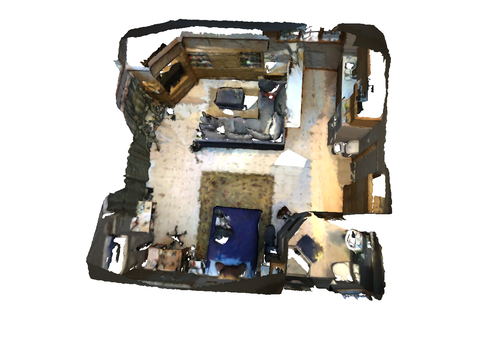} &
        \includegraphics[width=\imgwidth, height=\imgheight, keepaspectratio=false]{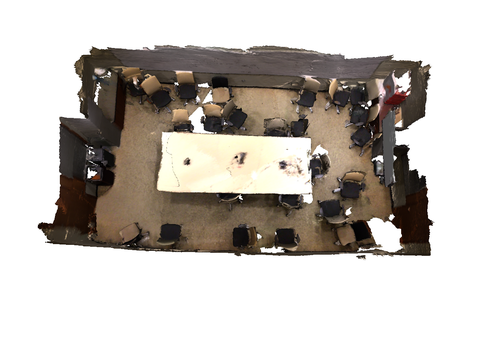} &
        \includegraphics[width=\imgwidth, height=\imgheight, keepaspectratio=false]{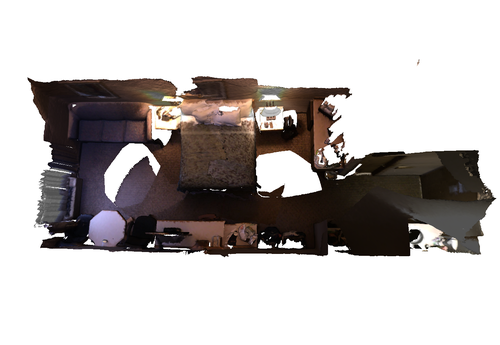} &
        \includegraphics[width=\imgwidth, height=\imgheight, keepaspectratio=false]{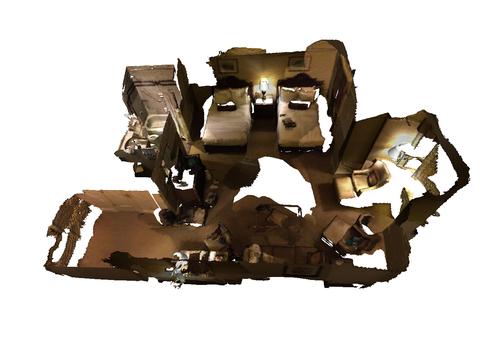} \\
        
        \rotatebox{90}{\hspace{11pt}Dr.Splat~\cite{JunSeong2025DrSD}} & 
        \includegraphics[width=\imgwidth, height=\imgheight, keepaspectratio=false]{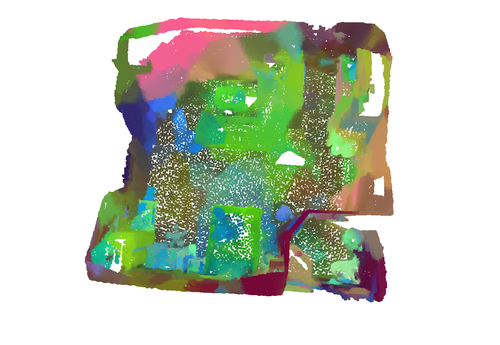} &
        \includegraphics[width=\imgwidth, height=\imgheight, keepaspectratio=false]{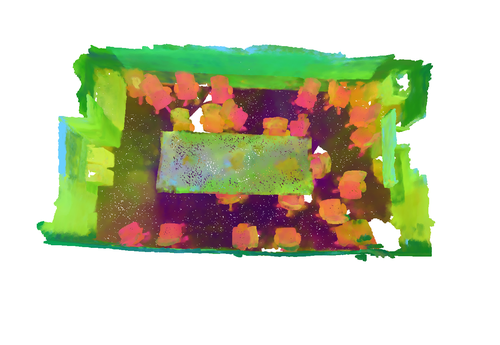} &
        \includegraphics[width=\imgwidth, height=\imgheight, keepaspectratio=false]{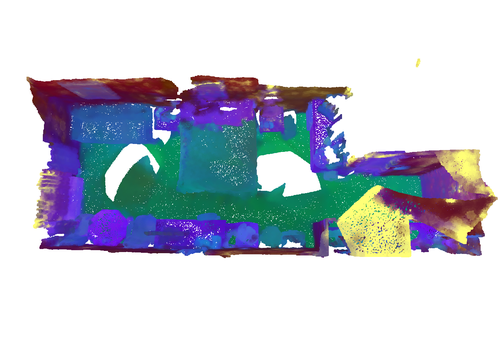} &
        \includegraphics[width=\imgwidth, height=\imgheight, keepaspectratio=false]{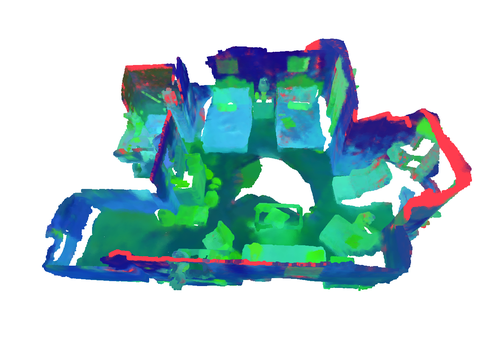} \\
        
        \rotatebox{90}{\hspace{22pt}Ours} & 
        \includegraphics[width=\imgwidth, height=\imgheight, keepaspectratio=false]{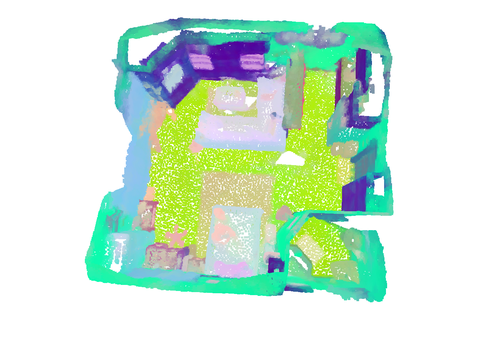} &
        \includegraphics[width=\imgwidth, height=\imgheight, keepaspectratio=false]{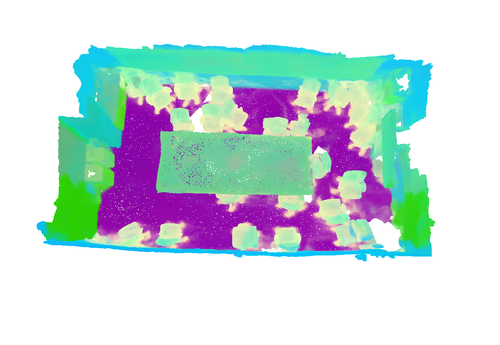} &
        \includegraphics[width=\imgwidth, height=\imgheight, keepaspectratio=false]{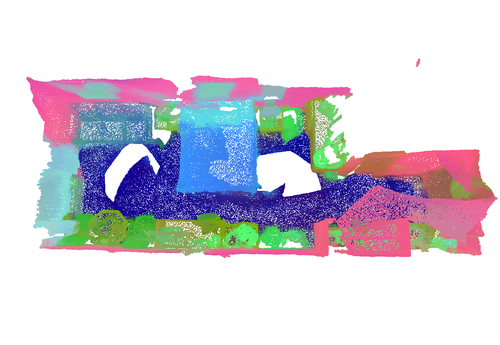} &
        \includegraphics[width=\imgwidth, height=\imgheight, keepaspectratio=false]{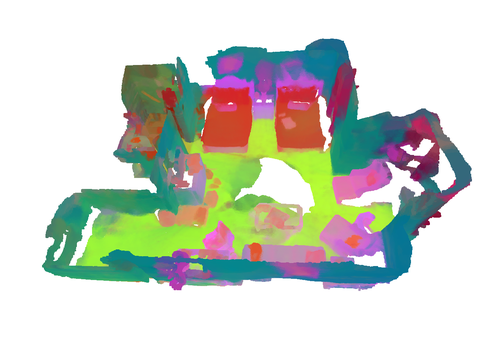} \\

    \end{tabular}
    \caption{\textbf{Qualitative Comparison on ScanNet.} Visual results of feature-transferred point clouds across diverse scenes. Compared to the baseline Dr.~Splat (middle row), our method (bottom row) significantly reduces semantic bleeding at object intersections with more accurate spatial boundaries.}

    \label{fig:scannet_pca_viz}
\end{figure*}

%% file: tabs/arch_ablation.tex
\begin{table}[h]
\centering
\resizebox{0.6\columnwidth}{!}{%
\begin{tabular}{lcccccc}
\cmidrule(l{1pt}r{1pt}){2-6}
 & \multicolumn{2}{c}{LERF (3D Retr.)} & & \multicolumn{2}{c}{ScanNet (PCD)} \\
\cmidrule{2-3} \cmidrule{5-6}
Configuration & mIoU & Acc@25 & & mIoU & mAcc \\
\midrule
Dr.~Splat~\cite{JunSeong2025DrSD} & 52.76 & \textbf{76.91} & & 31.66 & 48.64 \\
\midrule
SVRaster (vanilla) & 50.80 & 70.63 & & 30.62 & 48.90 \\
\quad + Mono. Sup. & 53.32 & 71.56 & & 31.37 & 49.78 \\
\quad + Mono. Sup. + Depth. Conf. & \textbf{53.37} & 72.91 & & \textbf{32.06} & \textbf{50.67} \\
\bottomrule
\end{tabular}%
}
\vspace{1mm}
\caption{Architecture ablation with SAM+CLIP features. }
\label{tab:arch_ablation}
\vspace{-20pt}
\end{table}

%% file: tabs/feature_ablation.tex
\begin{table}[h]
\centering
\resizebox{0.6\columnwidth}{!}{%
\begin{tabular}{llcccc}
\cmidrule(l{1pt}r{1pt}){3-6}
 & & \multicolumn{2}{c}{LERF (3D Retr.)} & \multicolumn{2}{c}{ScanNet (PCD)} \\
\cmidrule(lr){3-4} \cmidrule(lr){5-6}
Method & Feature & mIoU & Acc@25 & mIoU & mAcc \\
\midrule
Dr.~Splat~\cite{JunSeong2025DrSD} & CLIP & 52.76 & 76.91 & 31.66 & 48.64 \\
LESV (Ours) & CLIP & 53.37 & 72.91 & 32.06 & 50.67 \\
\midrule
Dr.~Splat~\cite{JunSeong2025DrSD} & AM-RADIO & 53.42 & 70.60 & 37.57 & 47.80 \\
LESV (Ours) & AM-RADIO & \textbf{56.11} & \textbf{85.21} & \textbf{53.22} & \textbf{70.41} \\
\bottomrule
\end{tabular}%
}
\vspace{1mm}
\caption{Disentangling representation vs.\ backbone features.}
\label{tab:feature_ablation}
\vspace{-20pt}
\end{table}

%% file: tabs/peak_memory.tex

\begin{table}[h]
\centering
\vspace{-3pt}
\resizebox{0.6\columnwidth}{!}{%
\begin{tabular}{lccccc}
\toprule
\multirow{2}{*}{ } & ramen  & figurines & teatime & waldo\_kitchen & \textbf{Average} \\
Num. Images & 131 & 299 & 177 & 187 & 198.5 \\
\midrule
Dr.~Splat~\cite{JunSeong2025DrSD} & 34 & 176 & 165 & 152 & 131.75 \\
\method~(ours) & \textbf{25} & \textbf{20} & \textbf{23} & \textbf{22} & \textbf{22.5} \\
\bottomrule
\end{tabular}
}
\vspace{1mm} 
\caption{Peak memory analysis.}
\label{tab:memory}
\vspace{-20pt}
\end{table}

%% file: tabs/speed.tex
\begin{table}[h]
    \centering
    \resizebox{0.6\linewidth}{!}{
    \begin{tabular}{l ccc}
        \toprule
        \scriptsize \textbf{Method} & \scriptsize \textbf{Preprocessing} & \scriptsize \textbf{Geometry} & \scriptsize \textbf{Feature Lifting} \\
        \midrule
        LangSplat \cite{Qin2023LangSplat3L} & $\sim$120 mins & $\sim$20 mins & $\sim$60 mins \\
        OpenGaussian \cite{Wu2024OpenGaussianTP} & $\sim$40 mins & $\sim$20 mins & $\sim$40 mins \\
        LaGa~\cite{cen2025tackling} (multi-level) & $\sim$120 mins & $\sim$20 mins & $\sim$70 mins \\
        Occam's LGS \cite{cheng2024occam} & $\sim$120 mins & $\sim$20 mins & \textbf{$\sim$3 mins} \\
        Dr. Splat \cite{JunSeong2025DrSD} & - & $\sim$20 mins & $\sim$10 mins \\
        \midrule
        \method~ (ours) & \textbf{$\sim$14 mins} & $\sim$15 mins & \textbf{$\sim$3 mins} \\
        \bottomrule
    \end{tabular}
    }
    \vspace{1mm}
    \caption{Computational efficiency comparison on a scene of $\sim$ 300 images.}
    \label{tab:speed_analysis}
    \vspace{-20pt}
\end{table}

%% file: secs/5_Conclusion.tex
\section{Conclusion}
In this paper, we presented \method, an efficient and accurate framework for open-vocabulary 3D scene understanding. By transitioning to an explicit Sparse Voxel Rasterization (SVRaster) backbone, we successfully mitigate the spatial ambiguity and semantic bleeding artifacts inherent in probabilistic 3DGS registration methods. Our methodology introduces two primary innovations for 3D language fields. First, by leveraging geometry grounded lifting with confidence gating rather than purely visual-guided feature lifting, we enable a robust, view-consistent and efficient feature registration framework, achieving an average feature lifting time of $\sim$3 minutes on the LERF dataset. Second, by directly lifting dense, language-aligned features from the AM-RADIO foundation model, we resolve complex part-to-whole semantic ambiguities without the prohibitive computational overhead of hierarchical 2D mask-pooling, reducing per-scene preprocessing time by over $\times8$ compared to LangSplat~\cite{Qin2023LangSplat3L}.

Empirical evaluations across the LERF and ScanNet benchmarks demonstrate that \method~establishes a new SoTA in open-vocabulary point cloud understanding, achieving a substantial gain of $+$21.56\% mIoU over the leading baseline. For 3D object retrieval, \method~achieves competitive performance against methods that rely on costly multi-level mask preprocessing, while reducing total processing time by 90\%. Furthermore, \method~naturally generalizes to 2D object retrieval and localization tasks, achieving performance comparable to SoTA methods optimized specifically for 2D.

\section{Limitation}
A primary bottleneck of \method~lies in the faithful upsampling of Vision Foundation Model (VFM) features. The effective resolution of our upsampled feature maps is currently limited to $256$ 
(cf. supplementary), which falls short of the spatial resolution provided by SAM masks. This resolution gap hinders performance on scenes containing small or densely packed objects, such as the \textit{figurines} scene. Developing methods for high-fidelity VFM feature upsampling that preserve semantic consistency remains an open and active research challenge. Furthermore, as with all direct registration methods, \method~does not support real-time feature rendering into novel views. While our deterministic fusion excels at 3D-native tasks and generalizes well to 2D projection, a more compact feature representation is needed to enable real-time rendering. Sparse coefficient fields, as proposed in LangSplat-v2~\cite{Li2025LangSplatV2H3}, represent a promising direction for future work.

%% file: secs/X_suppl.tex
\clearpage
\setcounter{page}{1}
\appendix

\section{Analysis of Deep Upsampling Degradation}
To isolate the impact of the upsampling strategy, we evaluate multiple configurations on the LERF 3D object retrieval benchmark, as detailed in Table \ref{tab:upsampling_analysis}. Notably, applying learning-based deep upsamplers such as AnyUp \cite{Wimmer2025AnyUpUF} yields strictly inferior performance compared to utilizing the raw, lower-resolution AM-RADIO features. Many of these architectures utilize cross-attention mechanisms where the final high-resolution output is technically a linear combination of the low-resolution value features. However, the attention weights themselves—computed via a highly non-linear softmax operation driven by high-resolution RGB queries—fundamentally alter the underlying feature distribution. 
\input{tabs/supp_upsample}

While these upsampled features showed that they can retain compatibility with their originally trained MLP heads ~\cite{Wimmer2025AnyUpUF}, our empirical results indicate they struggle significantly in zero-shot open-vocabulary tasks that rely more on raw cosine similarity. Specifically, this attention-based spatial mixing tends to homogenize the feature space, often yielding uniformly similarity scores across the image~\cite{Seo2025UpsampleAA}.
Consequently, while these upsamplers produce visually plausible pca feature maps, the resulting over-smoothing degrades the precise cosine similarity thresholding necessary for accurate 3D mask generation. Furthermore, we hypothesize this homogenization exacerbates spatial ambiguity when projecting these 2D features into a unified 3D voxel volume.

\section{Multi-Level TSDF Fusion for Adaptive Voxels}

When extracting meshes from adaptive sparse voxels via Truncated Signed Distance Function (TSDF) fusion, an important trade-off emerges between spatial resolution and geometric completeness. Fine-level voxels capture high-frequency details but inherently suffer from sparsity; many grid points remain unobserved yielding undefined values that manifest as holes in the final mesh. Conversely, coarse-level voxels provide more complete surfaces at the expense of fine detail. To achieve the best of both, we propose an adaptive, confidence-weighted blending mechanism. We first execute TSDF fusion at the finest octree level to preserve maximal detail. Subsequently, we iteratively fill low-confidence and unobserved regions by blending in TSDF values from progressively coarser levels. The complete multi-level progressive aggregation process is summarized in Algorithm \ref{alg:tsdf_fusion}. In Figure \ref{fig:ml_tsdf}, we provide a qualitative comparison between meshes extracted from sparse voxel grids clamped at different voxel levels versus our proposed multi-level TSDF fusion. 

\input{figs/ml_tsdf}

Furthermore, by directly leveraging the pre-optimized adaptive sparse volume rather than reconstructing a dense global grid, we circumvent extensive memory and computational overheads. The entire process of multi-level TSDF aggregation and subsequent mesh extraction requires less than 5 seconds for a standard scene containing approximately 300 images.

\section{Performance Across 2D and 3D Benchmarks}

To supplement the radar chart presented in our teaser Figure, Table \ref{tab:unified_benchmarks} provides a unified quantitative comparison across the ScanNet~\cite{Dai2017ScanNetR3} open-vocabulary 3D point cloud understanding benchmark (19 classes) and the LERF~\cite{Kerr2023LERFLE} open-vocabulary 2D and 3D object retrieval tasks. These results demonstrate that \method~maintains robust, state-of-the-art accuracy across all evaluated modalities. Since direct registration methods usually bypass rendering high-dimensional features, we adapt the 2D object retrieval evaluation for Dr.~Splat~\cite{JunSeong2025DrSD} and \method~by first computing the relevance scores directly in 3D for all Gaussians or voxels. We then render these scalar relevance values into 2D views for mask extraction, strictly following the standard LangSplat~\cite{Qin2023LangSplat3L} evaluation protocol.

\input{tabs/teaser_compare}
\input{algos/multi_level_tsdf}

\section{Performance on ScanNet200 Benchmark}
\input{tabs/scannet200}
We extend in Table~\ref{tab:scannet200} our evaluation to the 200-class ScanNet200 benchmark to assess open-vocabulary point cloud understanding. \method~maintains a substantial lead over Dr.~Splat~\cite{JunSeong2025DrSD}, achieving significant improvements of \textbf{+12.77\% mIoU} and \textbf{+16.39\% mAcc}. While OpenVoxel~\cite{huang2026openvoxel} also build on SVRaster, it addresses different objectives: OpenVoxel clusters voxels into instances via MLLMs for referring expression segmentation, whereas \method~registers dense embeddings per voxel for direct 3D queries at multi-level granularity, achieving stronger results on ScanNet-19 (53.22 vs.\ 31.6 mIoU). Notably, OpenVoxel shows ${\sim}$10\% mIoU sensitivity to SAM mask quality on 3D referring segmentation task~\cite{huang2026openvoxel}, confirming the intuition behind \method's design: bypassing mask-level dependencies with dense features is a promising direction for language-based 3D scene understanding.

\section{Real-World Applications in Autonomous Driving}
To demonstrate the scalability and practical utility of \method~in complex, unbounded environments, we extend experiment to real-world autonomous driving scenarios.
As illustrated in Figure \ref{fig:kitti_seg}, we deploy our pipeline to perform open-vocabulary 3D segmentation on KITTI-360~\cite{Liao2021KITTI360AN} with key driving semantics, including "road", "sidewalk". Our method successfully extracts accurate 3D masks, proving its viability for downstream trajectory planning or robust scene understanding.

Furthermore, the spatial grounding of \method~natively facilitates 3D scene manipulation. In Figure \ref{fig:scene_edit}, we showcase an open-vocabulary 3D scene editing application. By querying the structured volume for "red car", we localize the object's constituent voxels. We then directly manipulate their underlying Spherical Harmonic (SH) coefficients, altering the vehicle's appearance from red to black in 3D space. This capability highlights the  potential of \method~as a robust, editable 3D foundation for autonomous driving simulation or automated data augmentation.
\input{figs/kitti_seg}
\input{figs/kitti_edit}

\section{Additional Qualitative Results on LERF}

In Figure \ref{fig:full_qualitative_lerf}, we provide extended qualitative comparisons of 3D open-vocabulary object retrieval on the LERF~\cite{Kerr2023LERFLE} benchmark. Consistent with our main text findings, unstructured 3D Gaussian Splatting baselines such as LangSplat~\cite{Qin2023LangSplat3L} and Dr.Splat~\cite{JunSeong2025DrSD} struggle with spatial ambiguity, frequently exhibiting fuzzy boundaries, ray-like spillovers, and incomplete mask generation. In contrast, \method~demonstrates sharp semantic boundaries, cleanly isolating both large structural elements (e.g., ``yellow desk'') and extremely small targets (e.g., ``miffy''). Furthermore, our method successfully differentiates between closely related concepts such as a ``coffee mug'' versus ``tea in a glass''.

\section{Additional Qualitative Results on ScanNet}

In Figure \ref{fig:supp_scannet_full_viz}, we provide extended qualitative evaluations on the ScanNet~\cite{Dai2017ScanNetR3} dataset to demonstrate more results in open-vocabulary point cloud understanding. We visualize both the Principal Component Analysis (PCA) projection of the high-dimensional feature space (top) and the corresponding open-vocabulary segmentation maps for the 19-class benchmark (bottom). The PCA visualizations reveal the representational bottlenecks of unstructured 3DGS baselines; Dr.~Splat~\cite{JunSeong2025DrSD} yields noisy, entangled feature distributions that cause semantic bleeding across adjacent geometries. Conversely, \method~produces cohesive, discriminative feature clusters with uniform responses within distinct objects. This feature-level superiority translates to the segmentation results with less semantics bleeding and more accurate boundaries.

\input{figs/supp_lerf_2d}
\input{figs/supp_scannet_v2}

\clearpage

%% file: tabs/supp_upsample.tex
\begin{table}[h]
    \centering
    \resizebox{0.8\linewidth}{!}{
    \begin{tabular}{l cc}
        \toprule
        \textbf{Configuration} & \textbf{mIoU (\%)} & \textbf{Acc@25 (\%)} \\
        \midrule
        Raw AM-RADIO (128) & 51.19 & 76.04 \\
        \midrule
        + Deep Upsampler (256) \cite{Wimmer2025AnyUpUF} & 48.02 & 72.77 \\
        + Sliding Window + Gaussian Weighting (256) & 49.79 & 79.47 \\
        \midrule
        \textbf{+ SCRA \& SCGA (Proposed \method)} & \textbf{56.11} & \textbf{85.21} \\
        \bottomrule
    \end{tabular}
    }
    \caption{Quantitative comparison of feature upsampling strategies on the LERF 3D object retrieval benchmark. Deep upsampling degrades performance compared to the proposed self-correcting sliding window approach.}
    \label{tab:upsampling_analysis}
    \vspace{-8mm}
\end{table}

%% file: figs/ml_tsdf.tex
\begin{figure*}[tb] 
    \centering
    \def\fgsize{0.32} 
    \scriptsize
    \setlength{\tabcolsep}{0.005\linewidth}
    \renewcommand{\arraystretch}{1.0}
    \begin{tabular}{cccc}
        
        & level 9 & level 13 & multi-level (ours) \\
        
        \rotatebox{90}{\hspace{45pt} Method} & 
        \includegraphics[clip=false, trim={0 0 0 0}, width=\fgsize\textwidth]{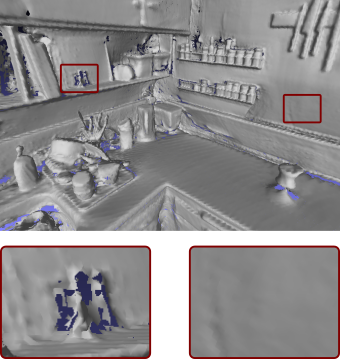} & 
        \includegraphics[clip=false, trim={0 0 0 0}, width=\fgsize\textwidth]{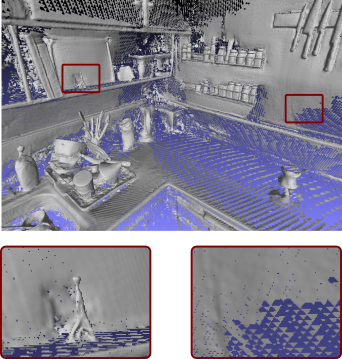} & 
        \includegraphics[clip=false, trim={0 0 0 0}, width=\fgsize\textwidth]{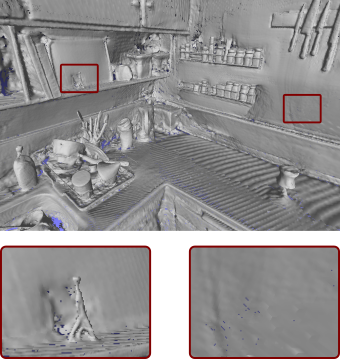} \\ 
        
    \end{tabular}
\caption{Qualitative comparison of mesh extraction strategies from adaptive sparse voxels. From left to right: clamping the grid to a coarse resolution (level 9), extracting at the finest resolution (level 13) which results in unobserved holes in blue, and extracted mesh from the proposed multi-level TSDF fusion.}
    \label{fig:ml_tsdf}
\end{figure*}

%% file: tabs/teaser_compare.tex
\begin{table}[t]
    \centering
    \resizebox{0.83\linewidth}{!}{
    \begin{tabular}{l ccc}
        \cmidrule(l{2pt}r{2pt}){2-4}
        & \multicolumn{2}{c}{\textbf{3D}} & {\textbf{2D}} \\
        \cmidrule(l{2pt}r{2pt}){2-3} \cmidrule(l{2pt}r{2pt}){4-4}
        {\hspace{7pt}} & \small 3D Obj. Retrieval {\hspace{7pt}} & \small 3D PCD Understanding {\hspace{7pt}} & \small 2D Obj. Retrieval \\
        Method & (mIoU \%) & (mIoU \%) & ~~~(mIoU \%) \\
        \midrule
        LangSplat~\cite{radford2021learning} & 10.97 & 3.78 & \underline{51.38} \\
        Dr.~Splat~\cite{JunSeong2025DrSD} & \underline{53.67} & 31.66 & 46.74 \\
        \midrule
        \method~(Ours) & \textbf{56.11} & \textbf{53.22} & \textbf{56.84} \\
        \bottomrule
    \end{tabular}
    }
    \vspace{1mm}
    \caption{Unified comparison across 2D and 3D benchmarks.}
    \label{tab:unified_benchmarks}
\end{table}

%% file: algos/multi_level_tsdf.tex
\begin{algorithm}[h]
\caption{Multi-Level TSDF Fusion via Confidence Blending}
\label{alg:tsdf_fusion}
\begin{algorithmic}[1]
\REQUIRE Finest TSDF $\Phi_{fine}$ and weights $\mathcal{W}_{fine}$, coarser levels $\mathcal{L}_{coarse}$, quantile $\tau_{q}$, temperature $T$.
\ENSURE Fused high-fidelity TSDF field $\Phi_{fused}$.
\FOR{each coarse level $l \in \mathcal{L}_{coarse}$}
    \STATE $\Phi_{coarse}, \mathcal{W}_{coarse}, \text{invmap} \leftarrow \text{ComputeCoarseTSDF}(l)$
    \STATE $\tau \leftarrow \text{Quantile}(\mathcal{W}_{fine}[\text{valid}], \tau_{q})$
    \FOR{each fine grid corner $c$ requiring update}
        \STATE $c_{coarse} \leftarrow \text{GetNearestCoarseCorner}(c, \text{invmap})$
        \IF{$\Phi_{fine}[c]$ is NaN}
            \STATE $\alpha \leftarrow 0$ \COMMENT{Trust coarse}
        \ELSIF{$\Phi_{coarse}[c_{coarse}]$ is NaN}
            \STATE $\alpha \leftarrow 1$ \COMMENT{Keep fine}
        \ELSE
            \STATE $\alpha \leftarrow \text{Sigmoid}\left(\frac{\mathcal{W}_{fine}[c] - \tau}{\tau \cdot T}\right)$
        \ENDIF
        \STATE $\Phi_{fine}[c] \leftarrow \alpha \cdot \Phi_{fine}[c] + (1 - \alpha) \cdot \Phi_{coarse}[c_{coarse}]$
    \ENDFOR
\ENDFOR
\RETURN $\Phi_{fine}$
\end{algorithmic}
\end{algorithm}

%% file: tabs/scannet200.tex
\begin{table}[h]
\centering
\resizebox{0.5\columnwidth}{!}{%
\begin{tabular}{lcccc}
            \toprule
            \multirow{2}{*}{\textbf{Method}} & \multicolumn{2}{c}{\textbf{ScanNet19}} & \multicolumn{2}{c}{\textbf{ScanNet200}} \\
            \cmidrule(lr){2-3} \cmidrule(lr){4-5}
            & mIoU & mAcc & mIoU & mAcc \\
            \midrule
            DR.~Splat~\cite{JunSeong2025DrSD} & 31.66 & 48.64 & 17.88 & 25.96 \\
            OpenVoxel~\cite{huang2026openvoxel} & 31.60 & 42.30 & - & - \\
            \method~(ours) & \textbf{53.22} & \textbf{70.41} & \textbf{30.65} & \textbf{42.35} \\
            \bottomrule
        \end{tabular}%
}
\vspace{1mm} 
\caption{Extended Comparison on ScanNet200.}
\label{tab:scannet200}
\end{table}

%% file: figs/kitti_seg.tex
\begin{figure}[htbp]
    \centering
    \def\fgsize{0.495} 
    \begin{subfigure}[b]{\fgsize\textwidth}
        \centering
        \includegraphics[width=\linewidth]{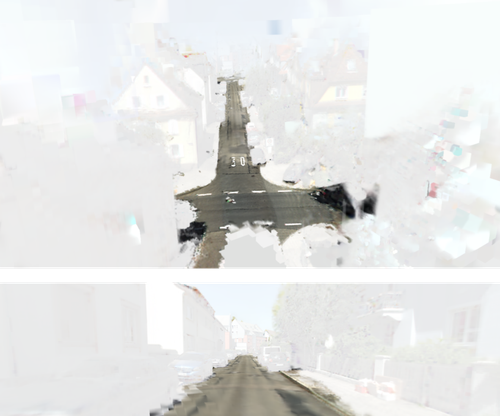}
        \caption{"road"}
        \label{fig:road}
    \end{subfigure}
    \hfill
    \begin{subfigure}[b]{\fgsize\textwidth}
        \centering
        \includegraphics[width=\linewidth]{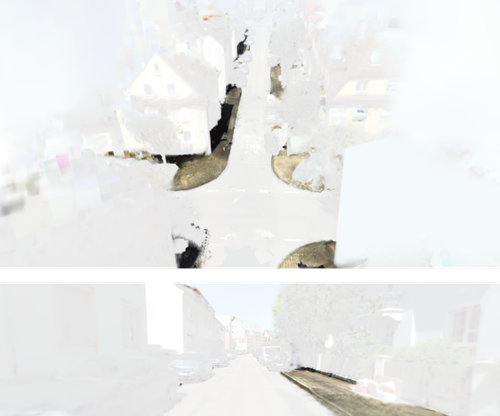}
        \caption{"side walk"}
        \label{fig:sidewalk}
    \end{subfigure}
    
    \vspace{1em} %
    
    \begin{subfigure}[b]{\fgsize\textwidth}
        \centering
        \includegraphics[width=\linewidth]{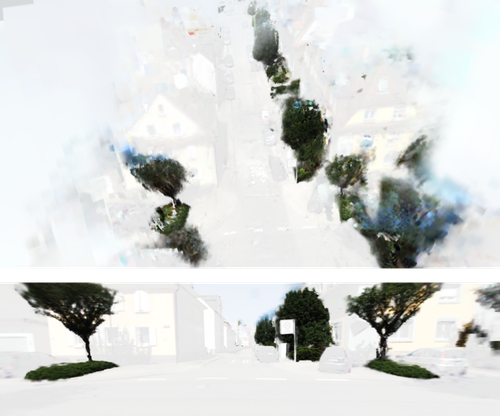}
        \caption{"vegetation"}
        \label{fig:vegetation}
    \end{subfigure}
    \hfill
    \begin{subfigure}[b]{\fgsize\textwidth}
        \centering
        \includegraphics[width=\linewidth]{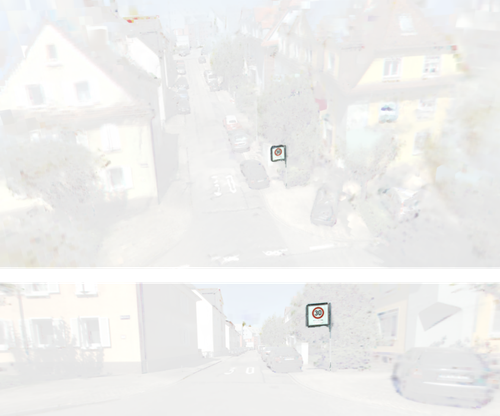}
        \caption{"traffic sign 30"}
        \label{fig:trafficsign}
    \end{subfigure}
    
    \caption{Open-vocabulary segmentation on KITTI-360~\cite{Liao2021KITTI360AN} for various text queries.}
    \label{fig:kitti_seg}
\end{figure}

%% file: figs/kitti_edit.tex
\begin{figure}[htbp]
    \centering
    
    \includegraphics[width=0.65
    \linewidth]{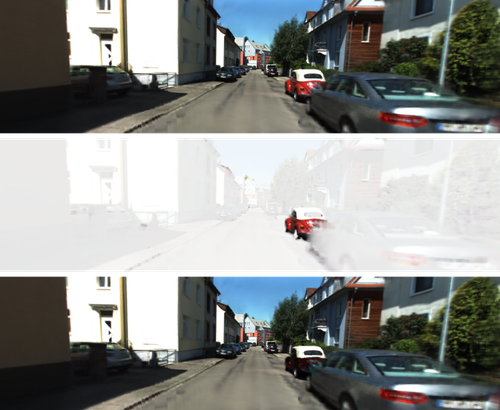} \\
    
    \caption{Scene editing example on the KITTI-360 dataset~\cite{Liao2021KITTI360AN}. From top to bottom: original rendering, open-vocabulary object retrieval on "red car", and the rendering after changing the color of such vehicle to black by modifying the SH coefficients.}
    \label{fig:scene_edit}
\end{figure}

%% file: figs/supp_lerf_2d.tex
\begin{figure*}[t]
    \centering
    \def\fgsize{0.235} 
    \scriptsize
    \setlength{\tabcolsep}{1pt} %
    \renewcommand{\arraystretch}{0.5} %
    
    \begin{subfigure}{\textwidth}
        \centering
        \begin{tabular}{ccccc}
            & ``bowl'' & ``cabinet'' & ``wavy noodles'' & ``yellow desk'' \\
            
            \rotatebox{90}{\hspace{5pt} LangSplat~\cite{Qin2023LangSplat3L}} & 
            \includegraphics[width=\fgsize\textwidth]{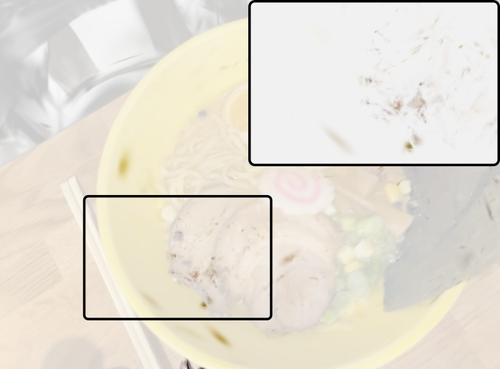} & 
            \includegraphics[width=\fgsize\textwidth]{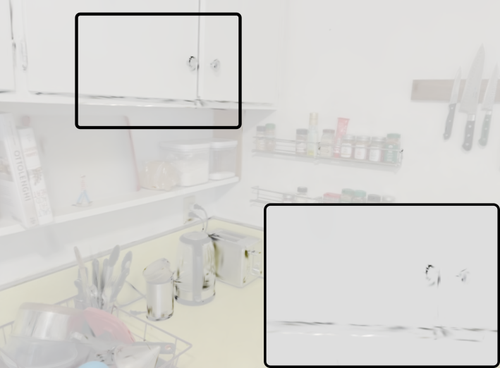} & 
            \includegraphics[width=\fgsize\textwidth]{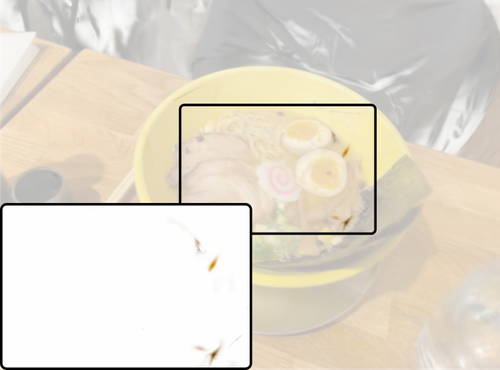} & 
            \includegraphics[width=\fgsize\textwidth]{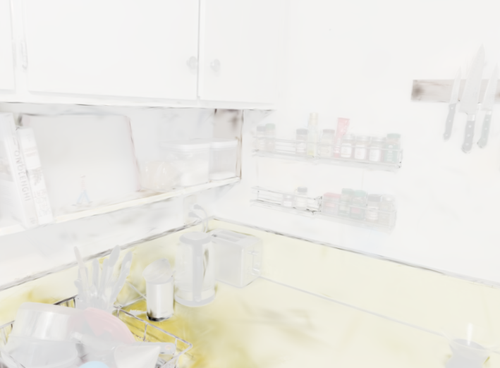} \\
            
            \rotatebox{90}{\hspace{7pt} Dr.Splat~\cite{JunSeong2025DrSD}} & 
            \includegraphics[width=\fgsize\textwidth]{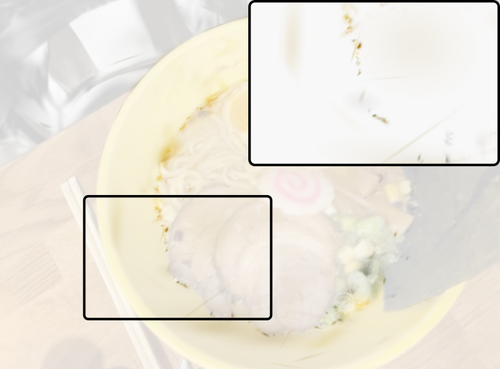} & 
            \includegraphics[width=\fgsize\textwidth]{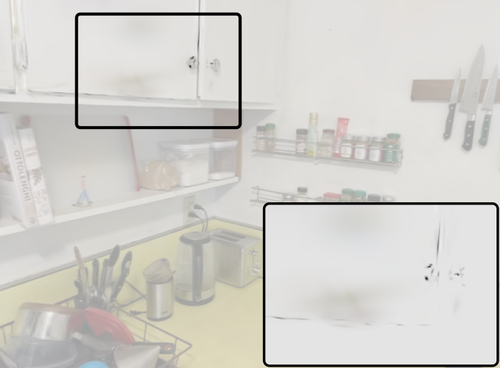} & 
            \includegraphics[width=\fgsize\textwidth]{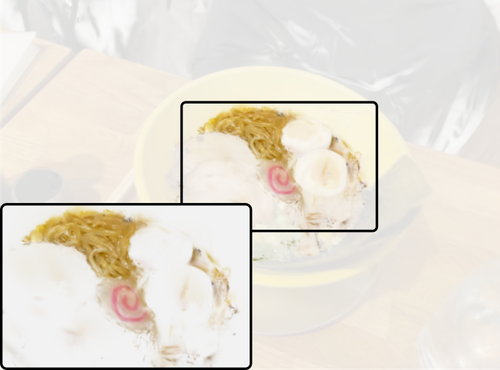} & 
            \includegraphics[width=\fgsize\textwidth]{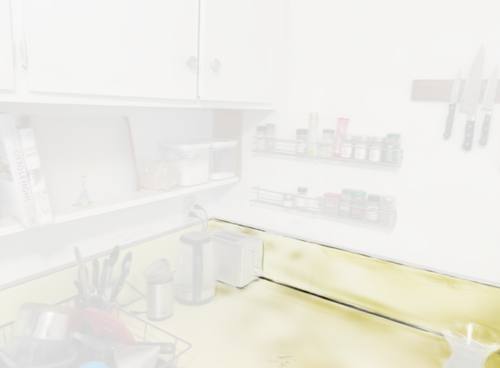} \\
            
            \rotatebox{90}{\hspace{6pt} \method~(Ours)} & 
            \includegraphics[width=\fgsize\textwidth]{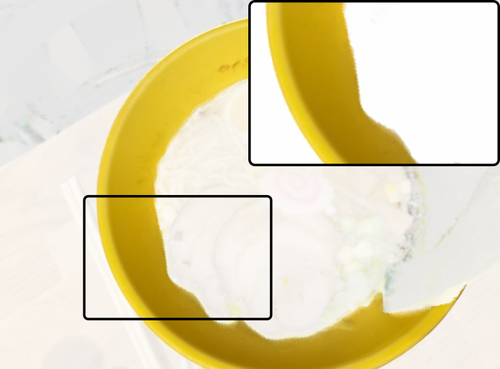} & 
            \includegraphics[width=\fgsize\textwidth]{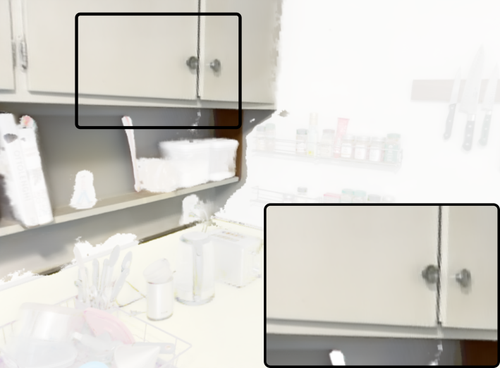} & 
            \includegraphics[width=\fgsize\textwidth]{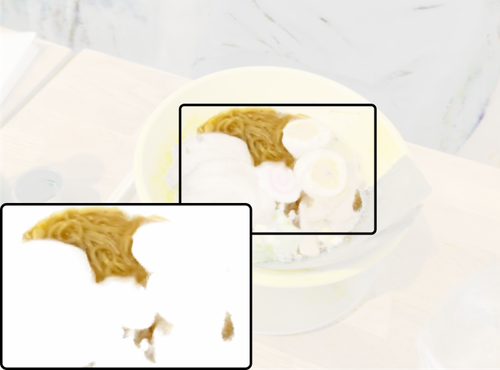} & 
            \includegraphics[width=\fgsize\textwidth]{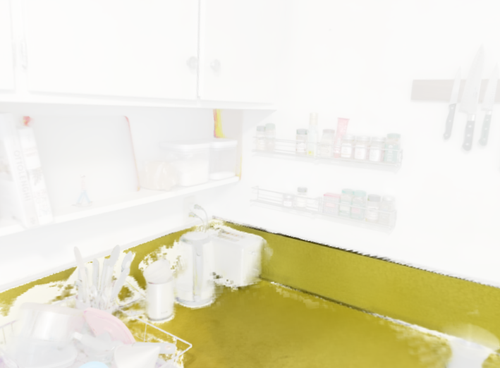} \\
        \end{tabular}
        \label{fig:qual_set1}
    \end{subfigure}
    
    \vspace{1em} %
    
    \begin{subfigure}{\textwidth}
        \centering
        \begin{tabular}{ccccc}
            & ``miffy'' & ``coffee mug'' & ``tea in a glass'' & ``coffee'' \\
            
            \rotatebox{90}{\hspace{5pt} LangSplat~\cite{Qin2023LangSplat3L}} & 
            \includegraphics[width=\fgsize\textwidth]{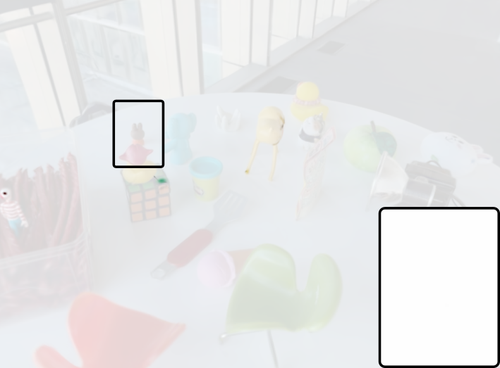} & 
            \includegraphics[width=\fgsize\textwidth]{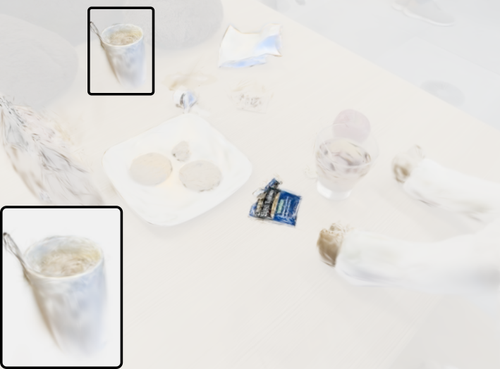} & 
            \includegraphics[width=\fgsize\textwidth]{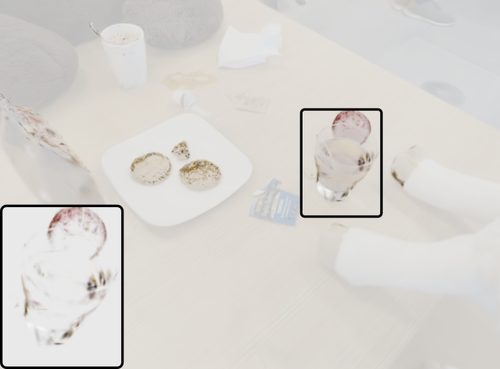} & 
            \includegraphics[width=\fgsize\textwidth]{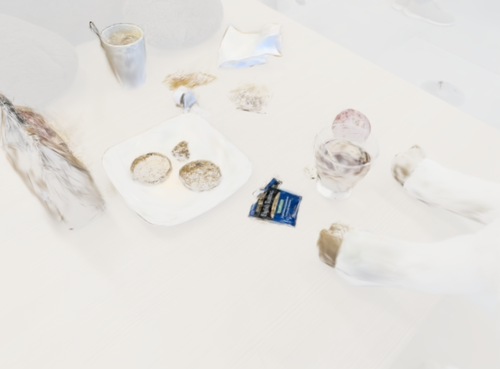} \\
            
            \rotatebox{90}{\hspace{7pt} Dr.Splat~\cite{li2025scenesplat}} & 
            \includegraphics[width=\fgsize\textwidth]{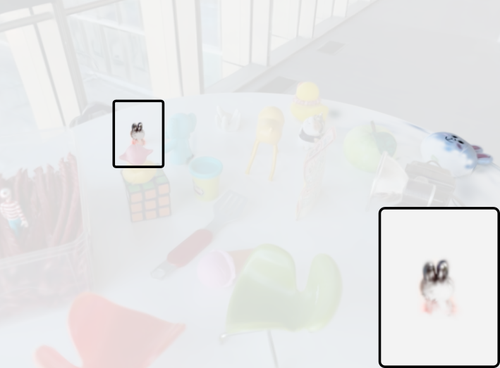} & 
            \includegraphics[width=\fgsize\textwidth]{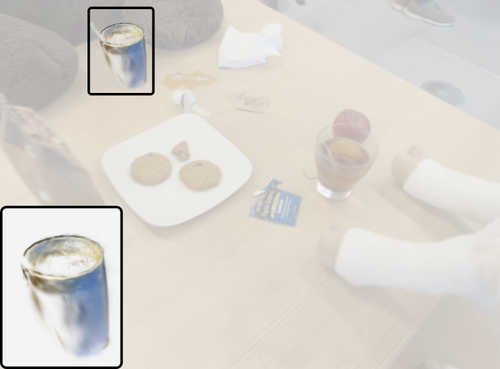} & 
            \includegraphics[width=\fgsize\textwidth]{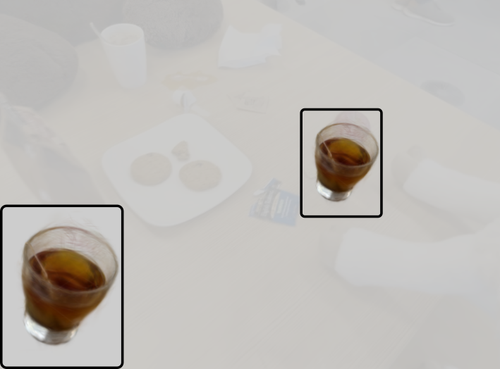} & 
            \includegraphics[width=\fgsize\textwidth]{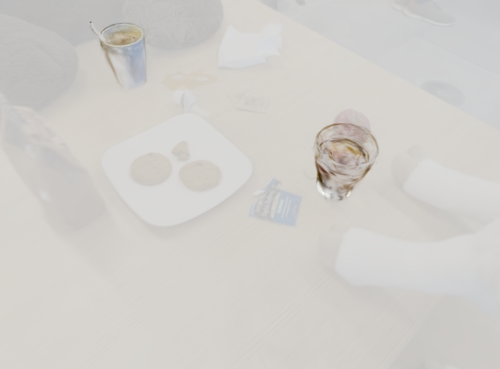} \\
            
            \rotatebox{90}{\hspace{6pt} \method~(Ours)} & 
            \includegraphics[width=\fgsize\textwidth]{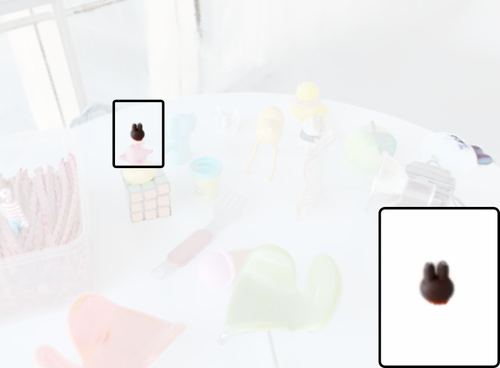} & 
            \includegraphics[width=\fgsize\textwidth]{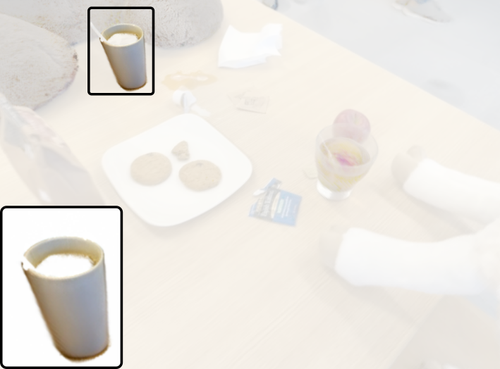} & 
            \includegraphics[width=\fgsize\textwidth]{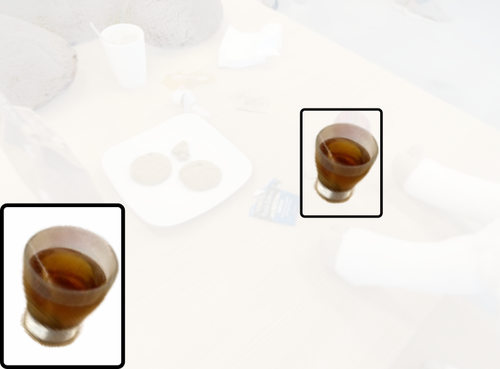} & 
            \includegraphics[width=\fgsize\textwidth]{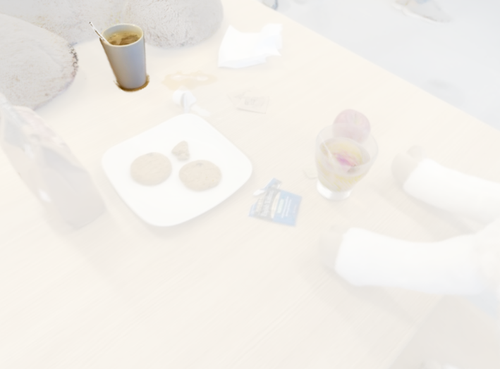} \\
        \end{tabular}
        \caption{Qualitative results for the second set of queries.}
        \label{fig:qual_set2}
    \end{subfigure}
    
    \caption{Qualitative comparison of 3D object retrieval on the LERF benchmark~\cite{Kerr2023LERFLE}. We compare our method against LangSplat~\cite{Qin2023LangSplat3L} and Dr.Splat~\cite{li2025scenesplat} across various text queries. Our method demonstrates sharper semantic boundaries and eliminates the spatial spillover artifacts common in unstructured 3DGS methods.}
    \label{fig:full_qualitative_lerf}
\end{figure*}

%% file: figs/supp_scannet_v2.tex
\begin{figure*}[t]
    \centering
    \newcommand{\imgwidth}{0.33\textwidth}
    \newcommand{\imgheight}{2.3cm} 
    \scriptsize
    \renewcommand{\arraystretch}{0.5}
    \setlength{\tabcolsep}{1pt}
    
    \begin{subfigure}{\textwidth}
        \centering
        \begin{tabular}{cccc}
            & Scene~0062 & Scene~0140 & Scene~0400 \\
            
            \rotatebox{90}{\hspace{16pt}RGB GT} &
            \includegraphics[width=\imgwidth, height=\imgheight, keepaspectratio=false]{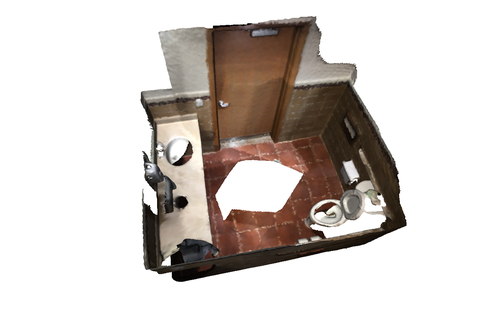} &
            \includegraphics[width=\imgwidth, height=\imgheight, keepaspectratio=false]{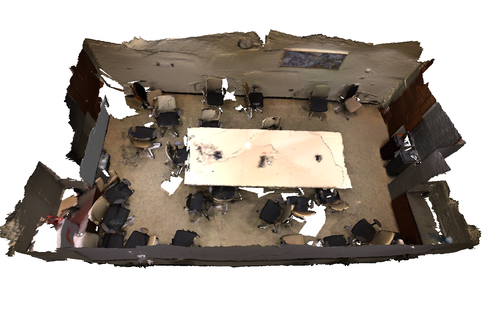} &
            \includegraphics[width=\imgwidth, height=\imgheight, keepaspectratio=false]{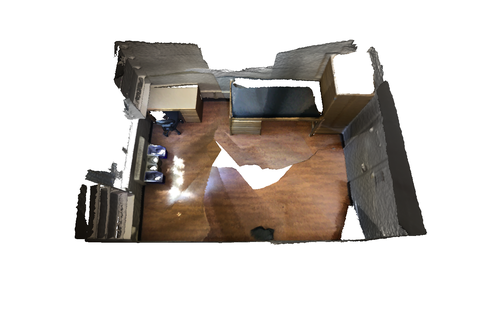} \\
            
            \rotatebox{90}{\hspace{11pt}Dr.Splat~\cite{JunSeong2025DrSD}} & 
            \includegraphics[width=\imgwidth, height=\imgheight, keepaspectratio=false]{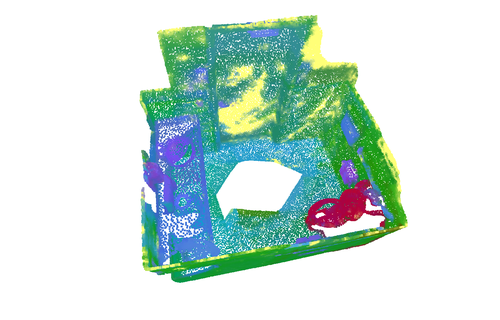} &
            \includegraphics[width=\imgwidth, height=\imgheight, keepaspectratio=false]{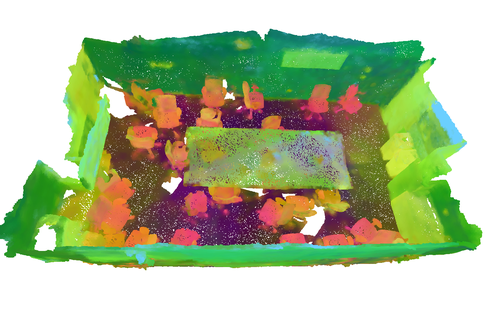} &
            \includegraphics[width=\imgwidth, height=\imgheight, keepaspectratio=false]{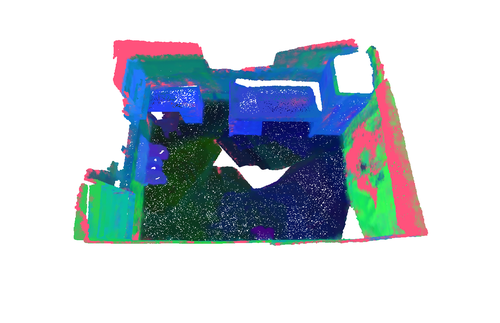} \\
            
            \rotatebox{90}{\hspace{7pt}\method~(Ours)} & 
            \includegraphics[width=\imgwidth, height=\imgheight, keepaspectratio=false]{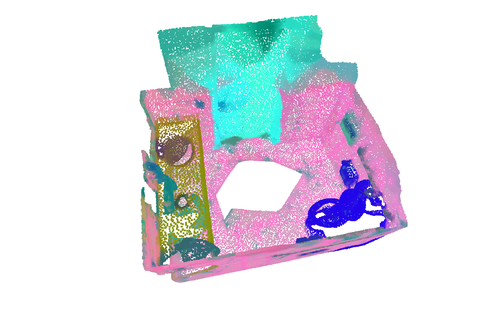} &
            \includegraphics[width=\imgwidth, height=\imgheight, keepaspectratio=false]{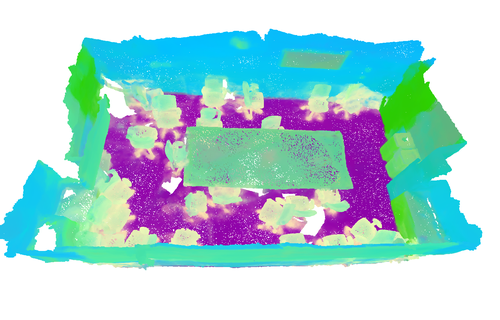} &
            \includegraphics[width=\imgwidth, height=\imgheight, keepaspectratio=false]{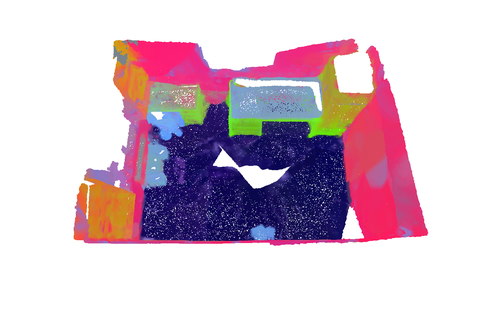} \\
        \end{tabular}
        \caption{PCA visualization of feature-transferred point clouds.}
        \label{fig:supp_scannet_pca_viz}
    \end{subfigure}
    
    \vspace{1em} %
    
    \begin{subfigure}{\textwidth}
        \centering
        \begin{tabular}{cccc}
            \rotatebox{90}{\hspace{12pt}Semantic GT} & 
            \includegraphics[width=\imgwidth, height=\imgheight, keepaspectratio=false]{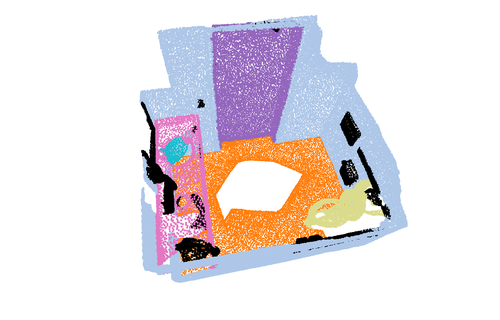} &
            \includegraphics[width=\imgwidth, height=\imgheight, keepaspectratio=false]{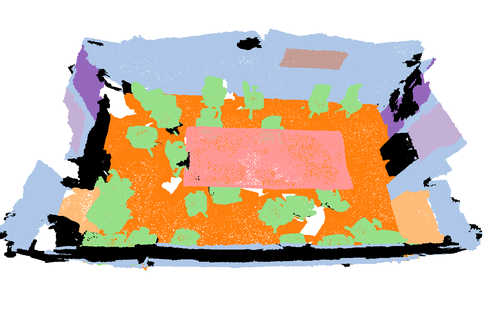} &
            \includegraphics[width=\imgwidth, height=\imgheight, keepaspectratio=false]{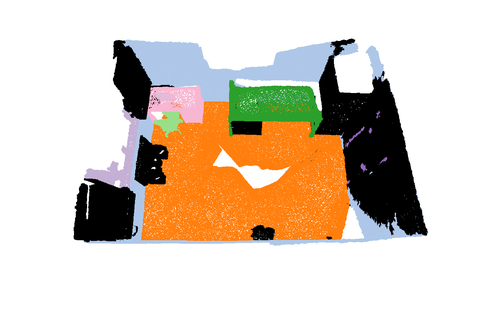} \\

            \rotatebox{90}{\hspace{11pt}Dr.Splat~\cite{JunSeong2025DrSD}} & 
            \includegraphics[width=\imgwidth, height=\imgheight, keepaspectratio=false]{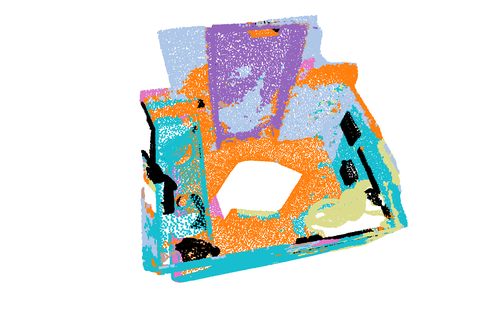} &
            \includegraphics[width=\imgwidth, height=\imgheight, keepaspectratio=false]{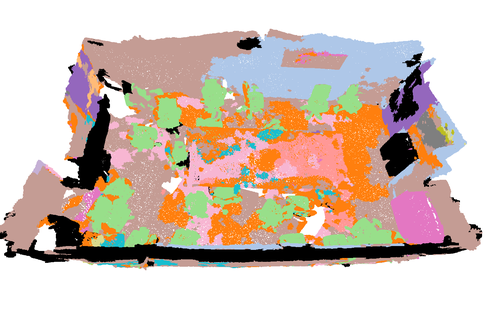} &
            \includegraphics[width=\imgwidth, height=\imgheight, keepaspectratio=false]{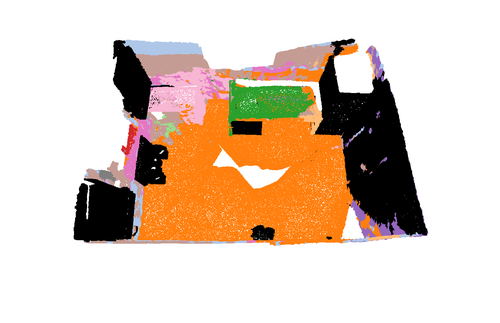} \\
            
            \rotatebox{90}{\hspace{7pt}\method~(Ours)} & 
            \includegraphics[width=\imgwidth, height=\imgheight, keepaspectratio=false]{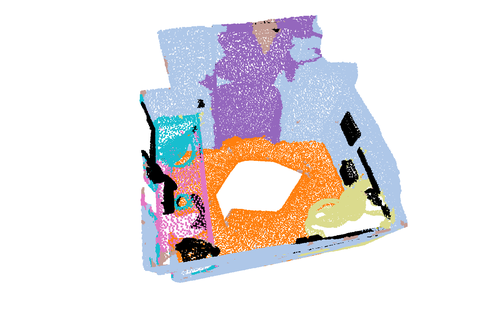} &
            \includegraphics[width=\imgwidth, height=\imgheight, keepaspectratio=false]{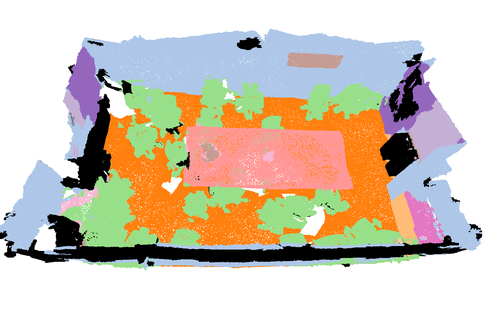} &
            \includegraphics[width=\imgwidth, height=\imgheight, keepaspectratio=false]{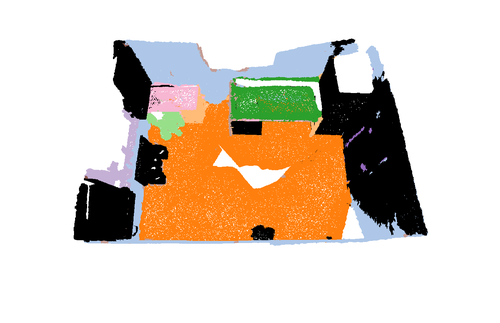} \\
        \end{tabular}
        \caption{Open-vocabulary point cloud segmentation results (19 classes).}
        \label{fig:supp_scannet_pcd_viz}
    \end{subfigure}
    
    \caption{\textbf{Qualitative Comparison on ScanNet.} (a) PCA visualizations representing the high-dimensional feature space. (b) Open-vocabulary segmentation maps. From top to bottom, groud truth from ScanNet~\cite{Dai2017ScanNetR3}, Dr.Splat~\cite{JunSeong2025DrSD}, and \method~results.}
    \label{fig:supp_scannet_full_viz}
\end{figure*}